\documentclass{article} % For LaTeX2e
\usepackage{iclr2026_conference,times}

\usepackage{amsmath,amsfonts,bm}

\def\eqref#1{equation~\ref{#1}}
\def\1{\bm{1}}

\DeclareMathAlphabet{\mathsfit}{\encodingdefault}{\sfdefault}{m}{sl}
\SetMathAlphabet{\mathsfit}{bold}{\encodingdefault}{\sfdefault}{bx}{n}

\usepackage{graphicx}
\usepackage{hyperref}
\usepackage{url}
\usepackage{multirow}
\usepackage{booktabs}
\usepackage{wrapfig}
\usepackage{caption}
\usepackage{subcaption}

\title{STORI: A Benchmark and Taxonomy for Stochastic Environments}

\author{%
  Aryan Amit Barsainyan \\
  \texttt{ophv91@visitor.nus.edu.sg} \\
  \texttt{aryan.barsainyan@gmail.com} \\
  \And
  Jing Yu Lim \\
  \texttt{jy\_lim@comp.nus.edu.sg} \\
  \And
  Dianbo Liu \\
  \texttt{dianbo@nus.edu.sg} \\
}

\iclrfinalcopy % Uncomment for camera-ready version, but NOT for submission.
\begin{document}

\maketitle

\begin{abstract}
Reinforcement learning (RL) techniques have achieved impressive performance on simulated benchmarks such as Atari100k, yet recent advances remain largely confined to simulation and show limited transfer to real-world domains. A central obstacle is environmental stochasticity, as real systems involve noisy observations, unpredictable dynamics, and non-stationary conditions that undermine the stability of current methods. Existing benchmarks rarely capture these uncertainties and favor simplified settings where algorithms can be tuned to succeed. The absence of a well-defined taxonomy of stochasticity further complicates evaluation, as robustness to one type of stochastic perturbation, such as sticky actions, does not guarantee robustness to other forms of uncertainty. To address this critical gap, we introduce STORI (STOchastic-ataRI), a benchmark that systematically incorporates diverse stochastic effects and enables rigorous evaluation of RL techniques under different forms of uncertainty. We propose a comprehensive five-type taxonomy of environmental stochasticity and demonstrate systematic vulnerabilities in state-of-the-art model-based RL algorithms through targeted evaluation of DreamerV3 and STORM. Our findings reveal that world models dramatically underestimate environmental variance, struggle with action corruption, and exhibit unreliable dynamics under partial observability. We release the code and benchmark publicly at \url{https://github.com/ARY2260/stori}, providing a unified framework for developing more robust RL systems.
\end{abstract}

\section{Introduction}
\label{introduction}

Reinforcement learning (RL) techniques have achieved impressive performance on simulated benchmarks such as Atari100k, yet recent advances remain largely confined to simulation and show limited transfer to real-world domains. A central obstacle is environmental stochasticity, as real systems involve noisy observations, unpredictable dynamics, and non-stationary conditions that undermine the stability of current methods \citep{antonoglou2022planning, 10.5555/3600270.3603094}. This challenge is especially acute for model-based RL, which must build world models to capture environment dynamics, a task that becomes significantly more complex when the environment exhibits multiple forms of uncertainty.

However, we lack a comprehensive stochastic environment benchmark that enables systematic development of RL methods robust to environmental stochasticity. Most widely used benchmarks, such as Atari games in the Arcade Learning Environment (ALE) \citep{bellemare13arcade}, are deterministic or nearly deterministic \citep{10.5555/3600270.3603094}. Although several approaches have introduced limited stochasticity, including sticky actions \citep{machado18arcade}, no-ops \citep{Mnih-V}, human starts \citep{nair2015massivelyparallelmethodsdeep}, and random frame skips \citep{brockman2016openaigym}---these modifications remain narrow in scope. To develop truly robust RL agents, we need benchmarks that systematically incorporate diverse forms of environmental uncertainty with granular control over both types and intensities of stochastic effects.

In this paper, we introduce STORI (STOchastic-ataRI), a benchmark that systematically incorporates diverse stochastic effects and enables rigorous evaluation of RL techniques under different forms of uncertainty. Alongside, we propose an updated taxonomy of stochasticity in RL environments, providing a unified framework for analyzing and comparing approaches. We leverage STORI to systematically investigate the reliability of world models under diverse forms of stochasticity and perform targeted evaluation probes to examine how learned dynamics respond to different stochastic challenges.

Our key contributions include:
\begin{itemize}
    \item \textbf{A comprehensive stochasticity taxonomy} with five distinct types: action-dependent noise, action-independent randomness, concept drift, representation learning challenges, and missing state information
    \item \textbf{STORI benchmark implementation} that systematically incorporates these stochasticity types into four Atari environments with granular parameter control  
    \item \textbf{Systematic evaluation} of state-of-the-art model-based RL algorithms (DreamerV3 and STORM) revealing fundamental vulnerabilities to environmental uncertainty
    \item \textbf{Targeted failure mode analysis} demonstrating that world models systematically underestimate variance, struggle with action corruption, and show unreliable dynamics under partial observability
    \item \textbf{Open-source framework} enabling researchers to develop and evaluate stochasticity-aware RL algorithms. We release the code and benchmark publicly at \url{https://github.com/ARY2260/stori}
\end{itemize}

\begin{figure}[htbp]
\centering
\begin{subfigure}[t]{\textwidth}
    \centering
    \includegraphics[width=0.9\textwidth]{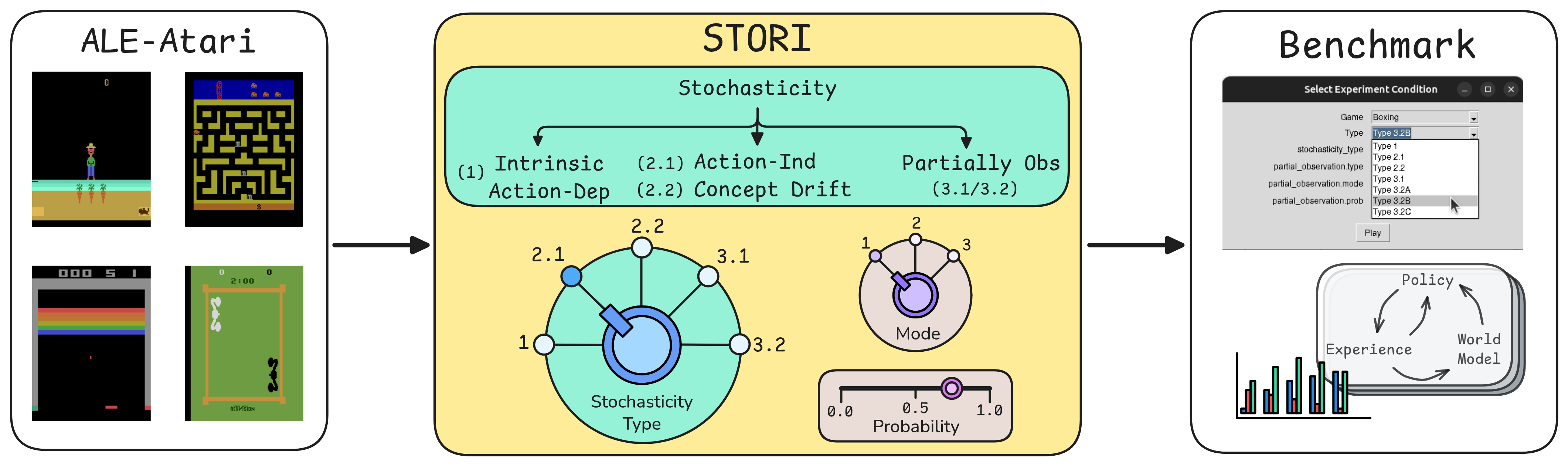}
    \caption{STORI benchmark framework}
    \label{fig:stori_overview}
\end{subfigure}
\vspace{0.3cm}
\begin{subfigure}[t]{0.9 \textwidth}
    \centering
    \includegraphics[width=\textwidth]{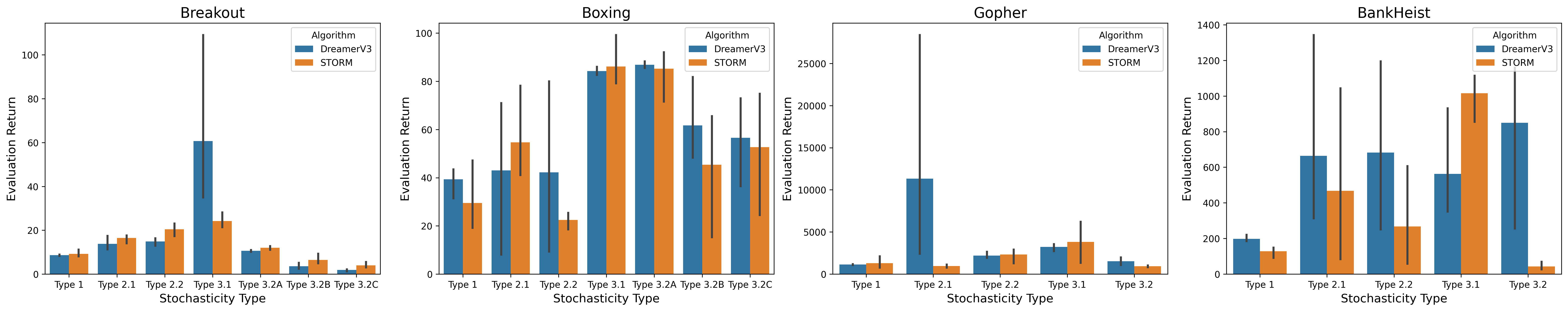}
    \caption{Performance across stochasticity types}
    \label{fig:performance_results}
\end{subfigure}
\vspace{-0.2cm}
\caption{STORI benchmark and results. (a) Framework for systematic stochasticity evaluation. (b) DreamerV3 and STORM performance degradation under uncertainty (Types: 1=action corruption, 2.1=random events, 2.2=concept drift, 3.1=default, 3.2=missing information).}
\label{fig:stori_combined}
\end{figure}
\vspace{-0.3cm}

\section{Related Works}
\label{related work}

\paragraph{Stochastic Environment Benchmarks}
Recent efforts have addressed limitations of deterministic RL benchmarks by incorporating stochastic perturbations. Robust-Gymnasium \citep{ICLR2025_fcc22e5b} provides a modular framework for robust evaluation across sixty robotics and control tasks, introducing observation-disruptors, action-disruptors, and environment-disruptors for systematic robustness assessment. Similarly, \cite{zouitine2024rrlsrobustreinforcement} introduced a benchmark extending Gymnasium-MuJoCo with six tasks capturing environmental shifts. STORI shares similar motivations but offers complementary contributions. First, it uses Atari games as a canonical testbed for discrete, high-dimensional decision-making. Second, STORI explicitly includes temporal non-stationarities such as concept drift, a critical yet underexplored aspect of real-world uncertainty. Beyond comprehensive benchmarks, several works target specific perturbation types. \cite{NEURIPS2020_f0eb6568} examined adversarial state perturbations, \cite{han2022what} studied multi-agent adversarial attacks, and \cite{park2025ogbenchbenchmarkingofflinegoalconditioned} analyzed offline goal-conditioned RL under perturbations. These reveal vulnerabilities to specific noise sources but lack unified robustness suites. Sandbox frameworks like MiniHack \citep{samvelyan2021minihack} allow custom stochastic environments.

\paragraph{Stochasticity Taxonomy}
Early RL literature formalized uncertainty through Partially Observable Markov Decision Processes (POMDPs) \citep{suttonBarto2018}, addressing partial observability and stochastic transitions. Recent works like \cite{Vamplew2022} propose broader stochastic MDP classifications for policy evaluation and learning. Beyond formal stochasticity, \cite{liu2023stateful} introduces environment heterogeneity concerning spatial layout and dynamics variations. \cite{Lu_2018} addresses temporal non-stationarity or concept drift, where target concepts change due to shifting hidden contexts. STORI's taxonomy integrates existing perspectives into a unified, practical framework explicitly designed for benchmarking, allowing systematic instantiation of stochasticity classes as configurable perturbations.

\paragraph{World Model Benchmarks}
Recent world modeling advances include DreamerV3 \citep{hafner2024masteringdiversedomainsworld}, IRIS \citep{iris2023}, STORM \citep{zhang2023storm}, TransDreamer \citep{chen2024transdreamerreinforcementlearningtransformer}, TWM \citep{robine2023transformerbased}, and DIAMOND \citep{alonso2024diffusionworldmodelingvisual}. These have been evaluated using Atari 100k \citep{ye2021masteringatarigameslimited}, BSuite \citep{osband2020bsuite}, Crafter \citep{hafner2021crafter}, and DMLab \citep{beattie2016deepmindlab}. STORI introduces a stochasticity-driven framework to analyze world model performance under environmental uncertainties.

% 1. world model 

% 2. model based RL 
% 3. Environmental stochsticity 
\section{Environment Stochasticity and our motivation}
\vspace{-0.3cm}

\subsection{Environment Stochasticity}
\vspace{-0.3cm}
RL environments can be categorized by their predictability. Deterministic environments have fully predictable outcomes, while stochastic environments introduce uncertainty requiring agents to handle outcome variability. Following \citet{Stochastic-systems}, stochasticity includes: \textbf{Stationary Stochastic (Objective)} with consistent statistical distributions (coin tosses); \textbf{Stationary Stochastic (Subjective)} based on personal beliefs (expert forecasts); \textbf{Non-Stationary Stochastic} with time-varying dynamics (traffic patterns); and \textbf{Illusory or Complex Uncertainty} where probabilities are unreliable (nuclear accidents).
\vspace{-0.2cm}
\subsection{Mathematical Taxonomy of Stochasticity}
\vspace{-0.1cm}
We formalize five key types using transition function $P(s'|s, a)$ where $s$ is the current state, $a$ is the action, and $s'$ is the next state.
\vspace{-0.3cm}
\subsubsection{Type 1: Intrinsic Action-Dependent Stochasticity}
Unreliable action channels corrupt intended action $a$ into executed action $\tilde{a}$.
\begin{equation}
    P_{AD}(s'|s, a) = \sum_{\tilde{a} \in \mathcal{A}} C(\tilde{a}|a) P(s'|s, \tilde{a})
\end{equation}
For sticky actions: $C(\tilde{a}|a) = (1 - \alpha)\mathbb{I}\{\tilde{a} = a\} + \alpha \cdot u(\tilde{a})$
\vspace{-0.2cm}
\subsubsection{Type 2.1: Intrinsic Action-Independent Stochasticity (Random)}
Exogenous random events independent of agent actions, with random variable $\xi \sim F(\xi)$:
\begin{equation}
    P_{AI}(s'|s, a) = \int P(s'|s, a, \xi) dF(\xi)
\end{equation}
\vspace{-0.4cm}
\subsubsection{Type 2.2: Intrinsic Action-Independent Stochasticity (Concept Drift)}
Time-dependent dynamics $P_t(s'|s, a)$ with drift magnitude:
\begin{equation}
    \text{Drift}(t, t+\Delta t) = \mathcal{D}(P_t(\cdot|s, a) || P_{t+\Delta t}(\cdot|s, a))
\end{equation}

\subsubsection{Type 3.1: Agent-Induced Stochasticity (Representation Learning)}
\vspace{-0.3cm}
Rich observations requiring representation learning where $I(S; O) \approx H(S)$. Belief states update via:
\begin{equation}
b_{t+1}(s') \propto O(o_{t+1}|s') \sum_{s \in \mathcal{S}} P(s'|s, a_t) b_t(s)
\end{equation}
State aliasing is resolvable through better feature extraction. \textit{Example:} Standard Atari RGB frames contain all game information but require learning to extract from pixels.
\vspace{-0.2cm}
\subsubsection{Type 3.2: Agent-Induced Stochasticity (Missing State Variables)}
\vspace{-0.3cm}
Critical state variables are completely omitted where $I(S; O) \ll H(S)$. Creates fundamental state aliasing $O(o|s_i) = O(o|s_j) = 1$ for $s_i \neq s_j$ that persists regardless of representational capacity. Same belief update as Type 3.1 but fundamentally limited.

Requires history tracking: $h_t = \{o_1, a_1, o_2, a_2, \ldots, o_t\}$

\textit{Examples:} Breakout with invisible ball regions; Boxing with hidden opponents; partial screen occlusion.

\vspace{-0.2cm}
\subsection{Challenges for World Model Learning and Model Based RL in Stochastic Environments}
\vspace{-0.3cm}
In this section, we analyze potential challenges for MBRL in different types of stochastic environments. A world model, denoted $\tilde{P}_{\theta}$, aims to learn the true transition dynamics from data. Each form of stochasticity introduces a distinct challenge that can cause a mismatch between $\tilde{P}_{\theta}$ and the true dynamics.
\vspace{-0.3cm}
\paragraph{Challenge from Type 1 Stochasticity}
The world model must learn not only the environment's response to actions, $P(s'|s, \tilde{a})$, but also the action channel itself, $C(\tilde{a}|s, a)$. If the model fails to account for the action channel, its predictions will be systematically biased. The prediction error is the divergence between the model's direct prediction and the true, action-corrupted dynamics: $\text{Error} = \mathcal{D}(P_{AD}(s'|s, a) || \tilde{P}_{\theta}(s'|s, a))$, where $P_{AD}$ represents the true dynamics and $\tilde{P}_{\theta}$ represents the model prediction. The model's ability to control the environment is limited by the \textbf{action channel capacity}, which can be measured by the mutual information $I(A; \tilde{A} | S)$. A low-capacity channel is fundamentally difficult to model and exploit.
\vspace{-0.2cm}
\paragraph{Challenge from Type 2.1 Stochasticity} This introduces irreducible \textbf{aleatoric uncertainty} into the environment. A deterministic world model will fail completely. A probabilistic world model must accurately capture the variance of the outcomes. The world model must learn a distribution over next states. The core challenge is to match the variance of this distribution to the true environmental variance, which is inherent and cannot be reduced with more data. The prediction error is tied to the model's ability to capture this spread: $\text{Aleatoric Error} = |\text{Var}_{s' \sim P_{AI}}[s'] - \text{Var}_{s' \sim \tilde{P}_{\theta}}[s']|$. The world model must avoid being overconfident in its predictions and instead represent the full range of possible outcomes.
\vspace{-0.2cm}
\paragraph{Challenge from Type 2.2 Stochasticity} Concept drift causes the world model's learned parameters $\theta$ to become outdated. A model trained on data from time $t$ will be inaccurate at time $t+\Delta t$. The prediction error grows over time as the environment drifts away from the data the model was trained on. The accumulated error is a function of the drift magnitude: $\text{Prediction Error}(t+\Delta t) \propto \mathcal{D}(P_t(\cdot|s, a) || \tilde{P}_{\theta}(\cdot|s, a))$, where $\tilde{P}_{\theta}$ was trained on data from distributions around time $t$. This forces the model to either continuously adapt its parameters or have a mechanism to detect and react to the drift.
\vspace{-0.2cm}
\paragraph{Challenge from Type 3.1 \& 3.2 Stochasticity} The world model cannot operate on the true state $s$ and must instead learn a latent state representation $z_t$ from a history of observations $o_{1:t}$. The primary challenge is \textbf{state aliasing}. The uncertainty a world model faces is not just the environment's true stochasticity, but also the aliasing-induced variance. The total variance in outcomes given an observation $o$ is decomposed as: $\text{Var}[s'|o, a] = \mathbb{E}_{s \sim b(s|o)}[\text{Var}[s'|s, a]] + \text{Var}_{s \sim b(s|o)}[\mathbb{E}[s'|s, a]]$, where the first term represents true aleatoric uncertainty and the second term represents aliasing-induced uncertainty. The world model's latent dynamics, $\tilde{P}_{\theta}(z'|z, a)$, must implicitly handle the second term, which is purely an artifact of perception. In Type 3.2 environments, where entire state variables are missing, this aliasing uncertainty can become overwhelmingly large, making it nearly impossible to form an accurate belief state and rendering long-term prediction unreliable.

\vspace{-0.2cm}
\section{STORI - a Benchmark of Stochastic Environments for RL}
\vspace{-0.2cm}
In this section, we describe in details the benchmark environments we built for different types of stochasticity based on Atari-Arcade learning environment. Atari games such as Breakout, Boxing, Gopher and BankHeist were modified to allow fine-grain control of these stochasticity. The taxonomy for stochasticity in STORI is an extension of the summary of classification of stochasticity according to \citet{antonoglou2022planning}.Table \ref{stochasticity-table} presents the taxonomy of stochasticity, listing each type, its corresponding subtype, and the associated ID. Each type is explained in the following sections.
%\begin{quote}
%\textit{"Many environments are inherently stochastic and may be poorly approximated by a deterministic model. Partially observed environments may also be perceived by the agent as stochastic, whenever aliased states cannot be disambiguated. Similarly, large and complex environments may appear stochastic to a small agent with finite capacity."}
%\end{quote}

\begin{table}[htbp]
\centering
\begin{minipage}{0.4\textwidth}
\centering
\caption{Environment stochasticity taxonomy}
\label{stochasticity-table}
\begin{tabular}{@{}cll@{}}
\toprule
\textbf{ID} & \textbf{Type} & \textbf{Sub Type} \\
\midrule
0 & Deterministic & NA \\
1 & Action Dependent & NA \\
2.1 & Action Independent & Random \\
2.2 & Action Independent & Concept Drift \\
3.1 & Partially Observed & Representation \\
3.2 & Partially Observed & Missing State \\
\bottomrule
\end{tabular}
\end{minipage}
\end{table}

\vspace{-0.2cm}
\subsection*{Atari - Arcade Learning Environment}
\vspace{-0.2cm}
The Arcade Learning Environment (ALE) provides a foundational framework for applying RL to Atari 2600 games \citep{bellemare13arcade}. Built on the Stella emulator and integrated with Gymnasium \citep{brockman2016openaigym}, ALE supports over a hundred games with extensive configurability including observation types (RGB, grayscale, RAM), action spaces, and stochasticity parameters like sticky actions \citep{machado18arcade}. The Atari 100K benchmark \citep{ye2021masteringatarigameslimited} evaluates sample efficiency by assessing agents after only 100,000 environment steps (approximately two hours of gameplay).

\begin{figure}[htbp]
\centering
\begin{minipage}{0.7\textwidth}
\centering
\begin{subfigure}[t]{0.38\textwidth}
    \centering
    \includegraphics[width=\textwidth]{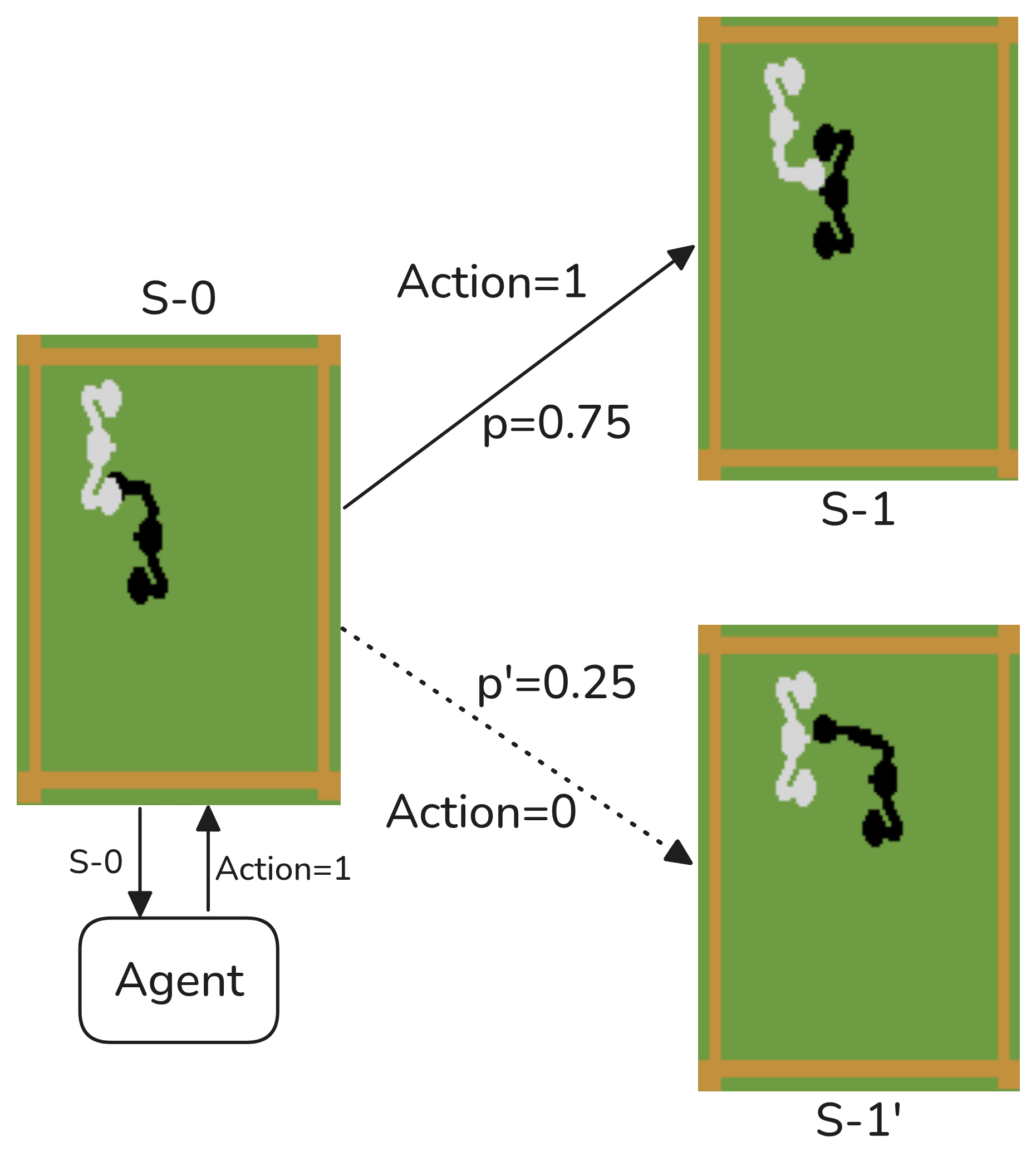}
    \caption{Type 1: Action-dependent}
    \label{fig:type1_boxing}
\end{subfigure}
\hfill
\begin{subfigure}[t]{0.48\textwidth}
    \centering
    \includegraphics[width=\textwidth]{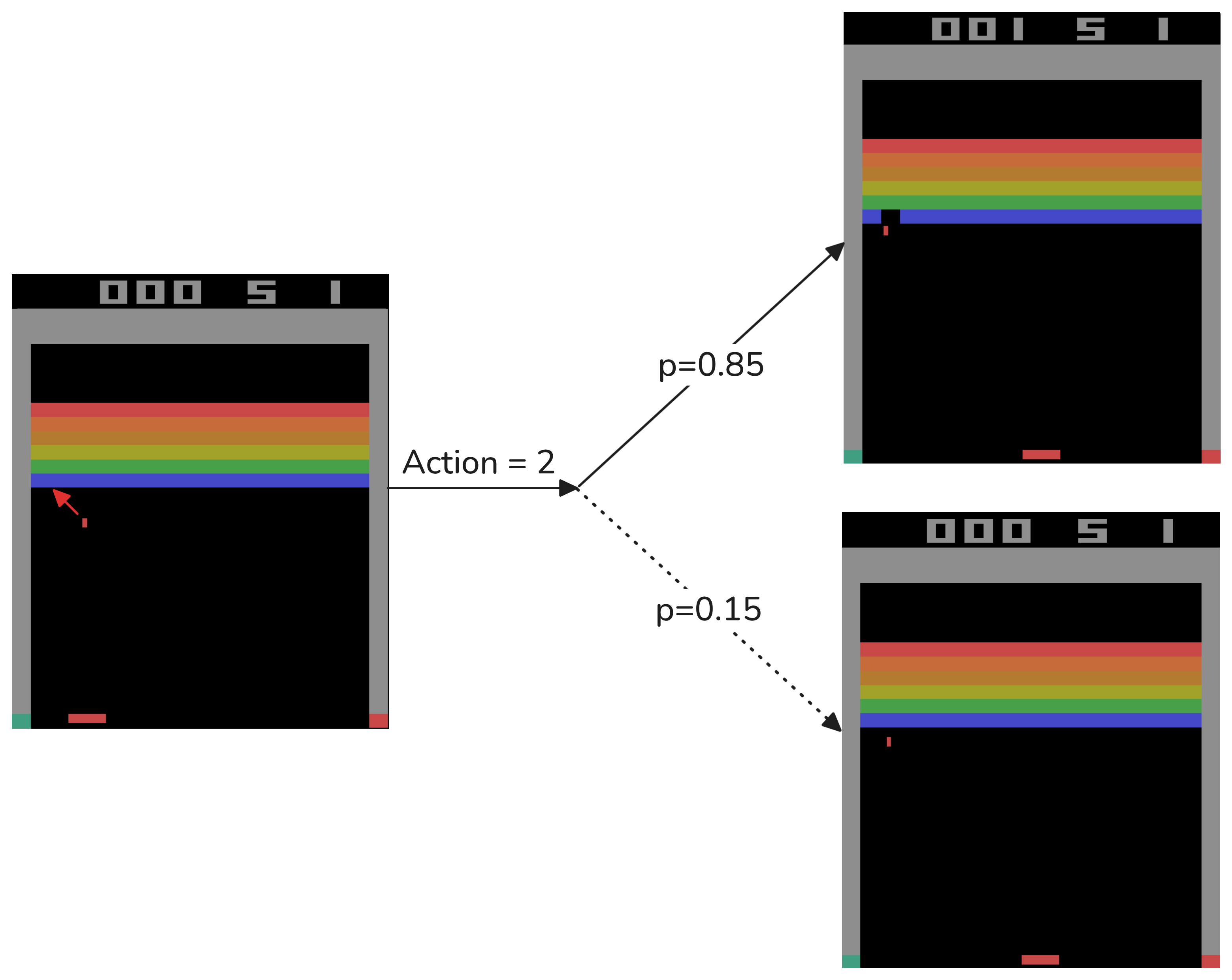}
    \caption{Type 2.1: Random events}
    \label{fig:type2_breakout}
\end{subfigure}

\vspace{0.1cm}

\begin{subfigure}[t]{0.48\textwidth}
    \centering
    \includegraphics[width=\textwidth]{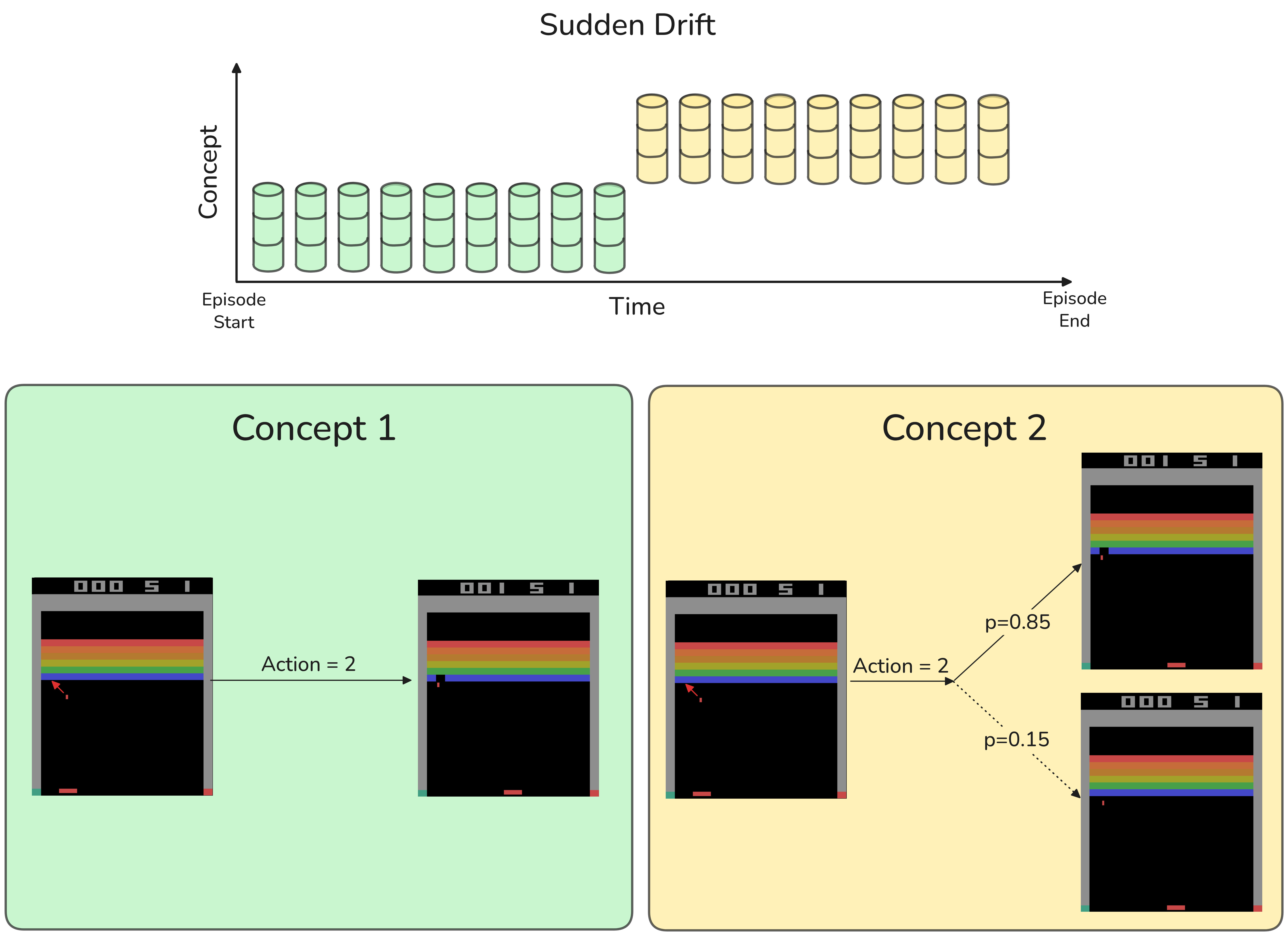}
    \caption{Type 2.2: Concept drift}
    \label{fig:concept_drift}
\end{subfigure}
\hfill
\begin{subfigure}[t]{0.48\textwidth}
    \centering
    \includegraphics[width=\textwidth]{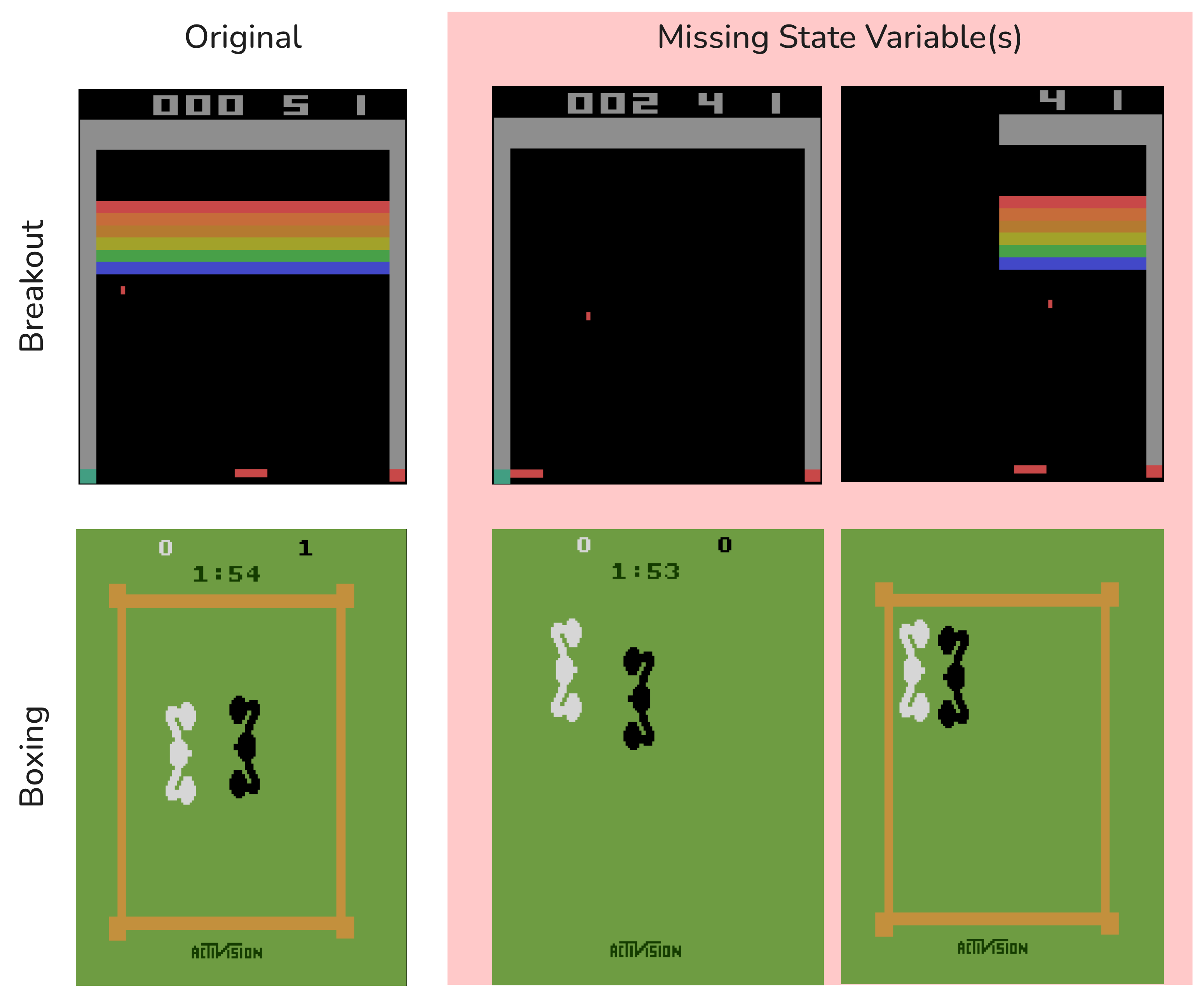}
    \caption{Type 3.2: Missing variables}
    \label{fig:partial_observability}
\end{subfigure}
\caption{Stochasticity types in STORI benchmark.}
\label{fig:stochasticity_examples}
\end{minipage}
\end{figure}

\vspace{-0.2cm}
\subsection*{Type 0: Deterministic Environment}
\vspace{-0.2cm}
Deterministic environments are those in which the next state is fully determined by the current state and the action taken. The state is completely observable and there is no randomness in the state transitions or rewards, meaning that the outcome of any action is predictable.

In the case of Atari, we consider the ground-truth labels of various state variables obtained directly from the RAM for each observation, following the approach of \citet{anand2020unsupervisedstaterepresentationlearning}. No additional stochasticity parameters are introduced, meaning that the environment is fully deterministic and corresponds to Type 0 in our taxonomy. Example for Breakout can be seen in figure \ref{fig:type0-breakout}.

\vspace{-0.2cm}
\subsection*{Type 1: Intrinsic Action Dependent Stochastic Environment}
\vspace{-0.2cm}
In environments with action-dependent intrinsic stochasticity, the environment may, by default, replace the agent’s chosen action with a random one. For instance, in $sticky\_action$ \citep{machado18arcade} scenarios, the environment can repeat the previous action with some probability. This results in varied outcomes even from the same state, with the stochastic effects limited to the state variables that can be influenced by the agent’s actions. An example of action-dependent intrinsic stochasticity with Atari Boxing can be seen in the \ref{fig:type1_boxing}.
\vspace{-0.2cm}

\subsection*{Type 2.1: Intrinsic Action Independent Stochastic Environment - Random}
\vspace{-0.2cm}
In action-independent random stochastic environments, randomness arises independently of the agent’s choices and affects state variables outside the agent’s direct control. This stochasticity, often due to external factors or inherent environmental noise, means that even with complete knowledge of the environment and carefully chosen actions, the next state cannot be predicted with certainty.

The figure \ref{fig:type2_breakout} illustrates an example of how this type of stochasticity can be modeled in Atari Breakout, where the paddle is moved to the right while the ball is on a trajectory to hit a block. In this case, there is a 0.15 probability that the ball bounces back without destroying the block or yielding any reward. Notably, this stochastic behavior is independent of the action of moving the paddle to the right.

% \begin{figure}[htbp]
% \begin{center}
% \includegraphics[width=0.7\textwidth]{figures/type-2-breakout.png}
% \end{center}
% \caption{The figure shows an example of type 2.1 stochasticity (Action Independent - Random) modeled in Atari Breakout, where the paddle is moved to the right while the ball is on a trajectory to hit a block. In this case, there is a 0.15 probability that the ball bounces back without destroying the block or yielding any reward.}
% \label{fig:type2-breakout}
% \end{figure}
\vspace{-0.2cm}
\subsection*{Type 2.2: Intrinsic Action Independent Stochastic Environment - Concept Drift
\vspace{-0.2cm}}
Environments with intrinsic action-independent concept drift can change over time independently of the agent's actions, a phenomenon known as concept drift \citep{Lu_2018}, which can generally be categorized into three types according to how the drift unfolds over time. In \textit{sudden drift}, the environment undergoes abrupt changes, forcing the agent to quickly adapt to new dynamics. In \textit{gradual or incremental drift}, the transition to new dynamics occurs slowly over time, requiring the agent to continuously adjust its policy. Finally, in \textit{recurring drift}, previously observed dynamics reappear in a cyclical or context-dependent manner, making long-term adaptation more challenging. Learning in such environments demands flexibility and the ability to detect and respond to changes.

In the case of Atari, most games have intrinsic incremental drift. As the agent levels up in the game, the difficulty of the game also increases. With a more fine-grain control over concept drift, other types of drift can also be achieved in Atari games as shown in figure \ref{fig:concept_drift}.

% \begin{table}[t]
% \caption{Types of Concept Drift}
% \label{tab:concept-drift}
% \begin{center}
% \begin{tabular}{ll}
% \multicolumn{1}{c}{\bf Drift Type} & \multicolumn{1}{c}{\bf Description} \\ \hline \\

% Sudden Drift & Environment changes abruptly. \\
% Gradual / Incremental Drift & Transition to new dynamics occurs slowly. \\
% Recurring Drift & Dynamics reappear in a cyclical or contextual manner. \\

% \end{tabular}
% \end{center}
% \end{table}

\vspace{-0.2cm}
\subsection*{Type 3.1: Partially Observed Environment (Representation Learning)}
\vspace{-0.2cm}
In partially observed environments, the agent does not have access to the full state information. When the state variables are represented differently from the true underlying environment, the agent must infer hidden information or learn an appropriate representation. This increases the complexity of decision making, since the agent must rely on approximate observations.

A typical example is the \textbf{Default Atari setting}, where the agent perceives only the screen image produced by the emulator after each action. These images are designed to approximate the true state, but they do not capture it fully. To enrich the observation, many implementations use a 4-frame skip with aggregation, which allows the agent to infer additional information, such as motion or rate of change over time, that is not apparent from a single frame.
\vspace{-0.2cm}
\subsection*{Type 3.2: Partially Observed Environment (Missing State Variable(s))}
\vspace{-0.2cm}
An important subclass of partially observed environments arises when information about certain state variables is missing, leaving the agent unable to observe critical aspects of the environment. This lack of information demands strategies that can manage uncertainty and make robust decisions despite gaps in perception. Such environments are common in real-world scenarios where sensors are limited, noisy, or unreliable.

Figure \ref{fig:partial_observability} illustrates type 3.2 environments using Atari Breakout and Boxing. In Breakout, examples include invisible blocks or a partially hidden screen, while in Boxing, examples include a hidden boxing ring or concealed clock and score information.

\vspace{-0.2cm}
\section{Experiments and Results}
\vspace{-0.2cm}

We evaluated DreamerV3 \citep{hafner2024masteringdiversedomainsworld} and STORM \citep{zhang2023storm} using STORI. DreamerV3 features a learned world model with actor-critic architecture, achieving robustness through fixed hyperparameters, normalization, and effective scaling. STORM employs a transformer backbone with stochastic variational modeling for strong sequence modeling and robustness. We focus on MBRL methods as model-free approaches require substantially more computational resources for meaningful results.

We selected four Atari 100K environments: Breakout, Boxing, Gopher (Agent-Optimal \citep{lim2025jedilatentendtoenddiffusion}), and BankHeist (Human-Optimal \citep{lim2025jedilatentendtoenddiffusion}). These provide action space diversity: Breakout (4 actions), Gopher (8), Boxing and BankHeist (18 each). Implementation details are in Appendix \ref{stori-implementation-appendix} and experiment stochasticity settings in Appendix \ref{exp-stochastic-modes}. Each algorithm was trained for 100K steps across baseline and modified environments for 3 seeds, evaluated on 100 episodes with mean return reported.

\subsection{Performance of DreamerV3 and STORM in Different Stochastic Environments}

\subsubsection{Breakout}
Stochasticity introduction caused marked performance decline versus default Type 3.1 environment (Figure \ref{fig:performance_results}), aligning with theoretical predictions. DreamerV3 initially outperformed STORM (60.71±41.89 vs 24.17±3.55) but STORM showed greater robustness across stochasticity types.

\begin{wraptable}{l}{0.45\textwidth}
\centering
\caption{Variance underestimation by world models.}
\begin{tabular}{@{}llr@{}}
\toprule
\textbf{Model} & \textbf{Type} & \textbf{Diff.} \\
\midrule
DreamerV3 & 3.1 & 1.25 \\
DreamerV3 & 2.1 & 300.34 \\
STORM & 3.1 & 1.32 \\
STORM & 2.1 & 325.21 \\
\bottomrule
\end{tabular}
\label{tab:variance_calibration}
\end{wraptable}
\vspace{-0.2cm}

Breakout's small action space (4 actions) creates high sensitivity to perturbations as incorrect LEFT/RIGHT actions immediately cause failure. Type 1 environments particularly impact control authority (Equation 14). Unlike Boxing, Breakout offers no recovery margin, amplifying uncertainty's long-term impact.

Type 2.1 environments caused severe struggles, with performance dropping to ~15\% of baseline, confirming that irreducible aleatoric uncertainty challenges deterministic world models. Type 2.2 concept drift (default→Type 3.2A after 300 steps) showed better performance than standalone Type 3.2A, suggesting adaptive mechanisms can leverage temporal structure.

\subsubsection{Boxing}
Boxing showed less pronounced performance decline. STORM initially led (86.18±11.29 vs 84.22±1.68) but DreamerV3 outperformed across several stochasticity types. Boxing's resilience stems from: (1) larger action space (18 actions) providing redundancy with functionally similar actions, and (2) recovery mechanisms through retreating/repositioning.

Type 3.2A (hidden score/clock) counterintuitively improved DreamerV3 performance (86.90±1.33 vs 84.22±1.68), suggesting non-essential information removal simplifies representation learning. For Type 3.2B (75\% right-half occlusion), agents showed adaptive behavior as they confined opponents to visible areas, transforming partial observability into strategic constraints.

Type 2.2 concept drift performed worse than standalone Type 3.2C, except DreamerV3's third seed learned to maximize early-episode scores before opponent invisibility, demonstrating strategic adaptation to predictable timing.

\subsubsection{Gopher and BankHeist}
Gopher showed high variability with Type 2.1 producing anomalous DreamerV3 performance (11,333.53±14,761.12) due to beneficial reward cancellation dynamics—suggesting implementation-specific edge case exploitation.

BankHeist exhibited divergent Type 3.2 outcomes: DreamerV3 achieved 849.80±514.15 while STORM fell to 43.10±22.85. DreamerV3 consistently adopted an unconventional policy, remaining near city gates and triggering inter-city transitions to loot nearby banks while avoiding stochastic modifications—effectively exploiting structural features to reduce tasks to near-deterministic subproblems rather than demonstrating genuine robustness. STORM explored broadly within cities, exposing itself to full stochastic impact. This highlights that high returns may reflect reward-maximizing shortcuts exploiting environment dynamics rather than genuine uncertainty resilience.
\subsection{Analysis of Error Types and World Model Failures}

To understand the specific failure modes of model-based RL under different stochasticity types, we conducted targeted analyses for each error category defined in Section 3.3.

% \subsubsection{Errors caused by type 1 stochasticity: Action Channel Mismatch}

% We analyzed model prediction errors by logging state transitions $(s_t, a_t, \tilde{a}_t, s_{t+1})$ in Type 1 environments and comparing model predictions $P_\theta(s'|s_t, a_t)$ against actual outcomes. The prediction error gap between correct executions ($a_t = \tilde{a}_t$) and corrupted actions ($a_t \neq \tilde{a}_t$) quantifies the action channel mismatch:

% $$\Delta = \text{Error}_{\text{corr}} - \text{Error}_{\text{ok}}$$

% Our analysis revealed that both DreamerV3 and STORM struggled to model the action corruption process, with $\Delta > 2.0$ in most cases, indicating systematic bias in world model predictions when actions are corrupted.

\subsubsection{Errors caused by type 2.1 stochasticity: Aleatoric Uncertainty and Concept Drift Analysis}

We ran a controlled probe of a single repeated action (action~3) for 1000 steps in both Type~3.1 (default setting,partially observed) and Type~2.1 (action-independent stochasticity) BankHeist environments, collecting states from the environment and predictions from the world models from DreamerV3 and STORM (Table \ref{tab:variance_calibration}). The resulting variance differences include:

\paragraph{Environment variance difference:}  
    \[
    \mathrm{Var}_{\text{env}}(\text{Type~2.1}) - \mathrm{Var}_{\text{env}}(\text{Type~3.1})
    \]  
    \(\rightarrow\) DreamerV3: \(299.097\), STORM: \(323.904\).  
    This confirms that Type~2.1 environments exhibit significantly higher true variance due to stochasticity.
    
 \paragraph{Model variance difference:}  
    \[
    \mathrm{Var}_{\text{model}}(\text{Type~2.1}) - \mathrm{Var}_{\text{model}}(\text{Type~3.1})
    \]  
    \(\rightarrow\) DreamerV3: \(0.00465\), STORM: \(0.01216\).  
    Both models predict nearly identical variance between environments despite the true variance increasing substantially.

Both DreamerV3 and STORM significantly underestimate the increased stochasticity present in Type~2.1 environments in BankHeist. While environment variance increases by approximately 300, the models’ predicted variances remain nearly constant. This mismatch highlights a lack of variance calibration under action-independent stochastic conditions, revealing a limitation in the world models’ ability to capture environment uncertainty accurately.

% We ran 1000 environment steps under a fixed action (action 3) for both Type~3.1 (default) and Type~2.1 (action-independent stochasticity) environments, collecting environment states and world model predictions. For each case, we compared the variance of true environment states $\mathrm{var}_{\text{env}}$ with the variance of predicted states $\mathrm{var}_{\text{model}}$ to assess variance calibration.

% For both DreamerV3 and STORM, the difference $\mathrm{var}_{\text{env}} - \mathrm{var}_{\text{model}}$ increased drastically in the Type~2.1 environment compared to Type~3.1:
% \begin{itemize}
%     \item \textbf{DreamerV3:} increased by $299.10$
%     \item \textbf{STORM:} increased by $323.90$
% \end{itemize}

% In contrast, the variance of the model predictions changed minimally between environments ($\Delta \mathrm{var}_{\text{model}} \approx 0.005$ for DreamerV3 and $\approx 0.012$ for STORM), indicating that both models failed to capture the increased true variance caused by the stochastic environment. These findings reveal a significant mismatch between predicted and actual variance under action-independent stochasticity, highlighting a limitation in world model calibration for stochastic transitions.

\begin{wraptable}{l}{0.5\textwidth}
\centering
\caption{Partial observability errors.}
\begin{tabular}{@{}lcc@{}}
\toprule
\textbf{Model} & \textbf{$\Delta$NLL} & \textbf{$\Delta$KL}\\
\midrule
DreamerV3 & 1.15±2.46 & $-0.18$±2.83 \\
STORM & $-3.32$±2.86 & 0.18±0.62 \\
\bottomrule
\end{tabular}
\label{tab:3_2_error_analysis}
\end{wraptable}
\vspace{-0.25cm}

For concept drift stochasticity, we measured the degradation ratio of model performance before and after the drift point. Table \ref{tab:drift_analysis} shows results for BankHeist Type 2.2.

The high degradation ratio for dynamics loss in DreamerV3 and STORM indicates that world model accuracy deteriorated significantly after the concept change, consistent with our theoretical prediction in Equation 12.

% state aliasing by finding state pairs $(s_A, s_B)$ that map to identical observations. The aliasing-induced uncertainty, measured as the variance in outcomes for the same observation, was consistently higher in Type 3.2 environments compared to Type 3.1, supporting our decomposition in Equation 13.

\subsubsection{Errors caused by type 3 stochasticity: State Aliasing Effects}

To investigate how well world models handle missing information, we designed a controlled experiment using BankHeist Type 3.2, where city blocks are randomly hidden in 75\% of observations. The key question: does a model's prediction accuracy depend on whether it can initially see the environment clearly?

\textbf{Experimental design:} We created six test scenarios and compared two starting conditions for each:
\begin{itemize}
    \item \textbf{Clear-start}: Model begins with city blocks visible, takes an action, observes the result
    \item \textbf{Obscured-start}: Model begins with city blocks hidden, takes the same action, sees the same result
\end{itemize}

\begin{figure}[htbp]
\centering
\begin{minipage}{0.7\textwidth}
\centering
\includegraphics[width=0.9\textwidth]{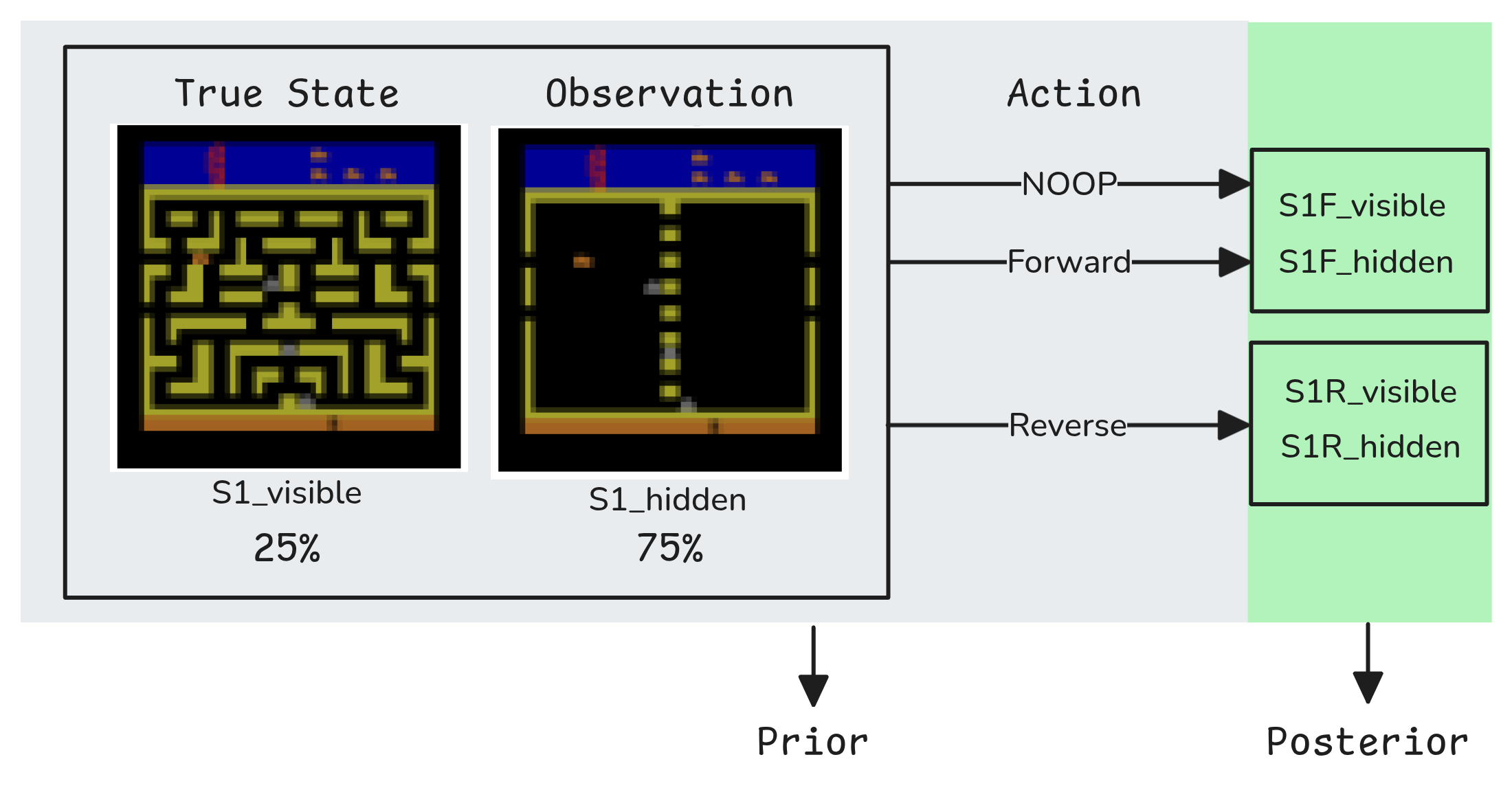}
\caption{Partial observability probe showing prediction errors when models start with clear vs. obscured observations.}
\label{fig:type_3_error_analysis}
\end{minipage}
\end{figure}

We then measured how ``surprised'' each model was by computing the difference in prediction error: $\Delta$NLL = NLL(obscured-start) $-$ NLL(clear-start) as in figure \ref{fig:type_3_error_analysis}.

\vspace{-0cm}
\begin{wraptable}{l}{0.4\textwidth}
\centering
\caption{Concept drift degradation.}
\begin{tabular}{@{}lcc@{}}
\toprule
\textbf{Model} & \textbf{Pre} & \textbf{Ratio} \\
\midrule
DreamerV3 & 5.4±15.8 & 6.94 \\
STORM & 4.5±2.4 & 5.04 \\
\bottomrule
\end{tabular}
\label{tab:drift_analysis}
\end{wraptable}
\vspace{-0cm}

\textbf{Key findings:} As in table \ref{tab:3_2_error_analysis}, DreamerV3 shows positive $\Delta$NLL values (1.15), meaning it makes significantly worse predictions when starting from obscured observations. In contrast, STORM shows negative $\Delta$NLL values ($-3.32$), indicating it actually performs slightly better when starting from limited information.

This reveals that DreamerV3's world model relies heavily on having complete initial observations to make accurate predictions. When city blocks are initially hidden, DreamerV3 struggles to maintain accurate beliefs about the environment state, requiring larger ``corrections'' to its internal model after seeing the action's outcome.

\textbf{Critical insight:} High task performance does not guarantee robust world model dynamics. Despite achieving strong returns in partially observable environments, DreamerV3's world model is more brittle when dealing with missing information compared to STORM.

\begin{wraptable}{l}{0.4\textwidth}
\centering
\caption{Model prediction errors under partial observability.}
\vspace{-0.2cm}
\begin{tabular}{@{}lcc@{}}
\toprule
\textbf{Model} & \textbf{$\Delta$NLL} & \textbf{$\Delta$KL}\\
\midrule
DreamerV3 & 1.15±2.46 & -0.18±2.83 \\
STORM & -3.32±2.86 & 0.18±0.62 \\
\bottomrule
\end{tabular}
\label{tab:3_2_error_analysis}
\end{wraptable}

\vspace{-0.3cm}
\section{Conclusions}

We introduced STORI, a systematic benchmark with a five-type taxonomy for evaluating RL algorithms under environmental stochasticity: action-dependent noise, action-independent randomness, concept drift, representation learning challenges, and missing state information.

Evaluation of DreamerV3 and STORM revealed systematic vulnerabilities in model-based approaches. Both algorithms struggle with action corruption, underestimate environmental variance by 300×, degrade 5-7× after concept drift, and show inconsistent reliability under partial observability. Strong task performance does not guarantee robust world model dynamics.

Limitations include cross-type comparison challenges, potential researcher bias in task selection, and computational constraints limiting evaluation to two algorithms. Future work should expand algorithmic coverage and develop stochasticity-aware world models. STORI provides a foundation for building more robust RL systems capable of handling real-world uncertainty.

\section{Reproducibility Statement}

We are committed to ensuring the reproducibility of our findings. All data, code, and implementation details necessary to replicate our experiments will be made available to the research community. Careful documentation accompanies the released resources to facilitate independent verification and reuse. The authors affirm that the results reported in this paper can be fully reproduced using the provided materials.

\section{Ethics Statement}
This work was conducted in accordance with established ethical standards for scientific research. All methods, analyses, and interpretations were carried out with a commitment to transparency, integrity, and responsible reporting. The authors confirm that no part of this research involved practices that could compromise fairness, safety, or the ethical treatment of data, participants, or systems.



\bibliography{iclr2026_conference}

\begin{thebibliography}{30}
\providecommand{\natexlab}[1]{#1}
\providecommand{\url}[1]{\texttt{#1}}
\expandafter\ifx\csname urlstyle\endcsname\relax
  \providecommand{\doi}[1]{doi: #1}\else
  \providecommand{\doi}{doi: \begingroup \urlstyle{rm}\Url}\fi

\bibitem[Alonso et~al.(2024)Alonso, Jelley, Micheli, Kanervisto, Storkey,
  Pearce, and Fleuret]{alonso2024diffusionworldmodelingvisual}
Eloi Alonso, Adam Jelley, Vincent Micheli, Anssi Kanervisto, Amos Storkey, Tim
  Pearce, and François Fleuret.
\newblock Diffusion for world modeling: Visual details matter in atari.
\newblock In \emph{Thirty-eighth Conference on Neural Information Processing
  Systems}, 2024.
\newblock URL \url{https://arxiv.org/abs/2405.12399}.

\bibitem[Anand et~al.(2020)Anand, Racah, Ozair, Bengio, Côté, and
  Hjelm]{anand2020unsupervisedstaterepresentationlearning}
Ankesh Anand, Evan Racah, Sherjil Ozair, Yoshua Bengio, Marc-Alexandre Côté,
  and R~Devon Hjelm.
\newblock Unsupervised state representation learning in atari, 2020.
\newblock URL \url{https://arxiv.org/abs/1906.08226}.

\bibitem[Antonoglou et~al.(2022)Antonoglou, Schrittwieser, Ozair, Hubert, and
  Silver]{antonoglou2022planning}
Ioannis Antonoglou, Julian Schrittwieser, Sherjil Ozair, Thomas~K Hubert, and
  David Silver.
\newblock Planning in stochastic environments with a learned model.
\newblock In \emph{International Conference on Learning Representations}, 2022.
\newblock URL \url{https://openreview.net/forum?id=X6D9bAHhBQ1}.

\bibitem[Beattie et~al.(2016)Beattie, Leibo, Teplyashin, Ward, Wainwright,
  Küttler, Lefrancq, Green, Valdés, Sadik, Schrittwieser, Anderson, York,
  Cant, Cain, Bolton, Gaffney, King, Hassabis, Legg, and
  Petersen]{beattie2016deepmindlab}
Charles Beattie, Joel~Z. Leibo, Denis Teplyashin, Tom Ward, Marcus Wainwright,
  Heinrich Küttler, Andrew Lefrancq, Simon Green, Víctor Valdés, Amir Sadik,
  Julian Schrittwieser, Keith Anderson, Sarah York, Max Cant, Adam Cain, Adrian
  Bolton, Stephen Gaffney, Helen King, Demis Hassabis, Shane Legg, and Stig
  Petersen.
\newblock Deepmind lab, 2016.
\newblock URL \url{https://arxiv.org/abs/1612.03801}.

\bibitem[{Bellemare} et~al.(2013){Bellemare}, {Naddaf}, {Veness}, and
  {Bowling}]{bellemare13arcade}
M.~G. {Bellemare}, Y.~{Naddaf}, J.~{Veness}, and M.~{Bowling}.
\newblock The arcade learning environment: An evaluation platform for general
  agents.
\newblock \emph{Journal of Artificial Intelligence Research}, 47:\penalty0
  253--279, jun 2013.

\bibitem[Brockman et~al.(2016)Brockman, Cheung, Pettersson, Schneider,
  Schulman, Tang, and Zaremba]{brockman2016openaigym}
Greg Brockman, Vicki Cheung, Ludwig Pettersson, Jonas Schneider, John Schulman,
  Jie Tang, and Wojciech Zaremba.
\newblock Openai gym, 2016.
\newblock URL \url{https://arxiv.org/abs/1606.01540}.

\bibitem[Chen et~al.(2024)Chen, Wu, Yoon, and
  Ahn]{chen2024transdreamerreinforcementlearningtransformer}
Chang Chen, Yi-Fu Wu, Jaesik Yoon, and Sungjin Ahn.
\newblock Transdreamer: Reinforcement learning with transformer world models,
  2024.
\newblock URL \url{https://arxiv.org/abs/2202.09481}.

\bibitem[Gu et~al.(2025)Gu, Shi, Wen, Jin, Mazumdar, Chi, Wierman, and
  Spanos]{ICLR2025_fcc22e5b}
Shangding Gu, Laixi Shi, Muning Wen, Ming Jin, Eric Mazumdar, Yuejie Chi, Adam
  Wierman, and Costas Spanos.
\newblock Robust gymnasium: A unified modular benchmark for robust
  reinforcement learning.
\newblock In Y.~Yue, A.~Garg, N.~Peng, F.~Sha, and R.~Yu (eds.),
  \emph{International Conference on Representation Learning}, volume 2025, pp.\
   102014--102041, 2025.
\newblock URL
  \url{https://proceedings.iclr.cc/paper_files/paper/2025/file/fcc22e5b7d5d2155d994da22d045f0a6-Paper-Conference.pdf}.

\bibitem[Hafner(2021)]{hafner2021crafter}
Danijar Hafner.
\newblock Benchmarking the spectrum of agent capabilities.
\newblock \emph{arXiv preprint arXiv:2109.06780}, 2021.

\bibitem[Hafner et~al.(2024)Hafner, Pasukonis, Ba, and
  Lillicrap]{hafner2024masteringdiversedomainsworld}
Danijar Hafner, Jurgis Pasukonis, Jimmy Ba, and Timothy Lillicrap.
\newblock Mastering diverse domains through world models, 2024.
\newblock URL \url{https://arxiv.org/abs/2301.04104}.

\bibitem[Han et~al.(2024)Han, Su, He, Han, Yang, and Miao]{han2022what}
Songyang Han, Sanbao Su, Sihong He, Shuo Han, Haizhao Yang, and Fei Miao.
\newblock What is the solution for state-adversarial multi-agent reinforcement
  learning?
\newblock \emph{Transactions on Machine Learning Research (TMLR)}, 2024.

\bibitem[Kumar \& Varaiya(1986)Kumar and Varaiya]{Stochastic-systems}
P.~R. Kumar and Pravin Varaiya.
\newblock \emph{Stochastic systems: estimation, identification and adaptive
  control}.
\newblock Prentice-Hall, Inc., USA, 1986.
\newblock ISBN 013846684X.

\bibitem[Lim et~al.(2025)Lim, Ikram, Yu, Ma, Leong, and
  Liu]{lim2025jedilatentendtoenddiffusion}
Jing~Yu Lim, Zarif Ikram, Samson Yu, Haozhe Ma, Tze-Yun Leong, and Dianbo Liu.
\newblock Jedi: Latent end-to-end diffusion mitigates agent-human performance
  asymmetry in model-based reinforcement learning, 2025.
\newblock URL \url{https://arxiv.org/abs/2505.19698}.

\bibitem[Liu et~al.(2023)Liu, Shah, Boussif, Meo, Goyal, Shu, Mozer, Heess, and
  Bengio]{liu2023stateful}
Dianbo Liu, Vedant Shah, Oussama Boussif, Cristian Meo, Anirudh Goyal, Tianmin
  Shu, Michael~Curtis Mozer, Nicolas Heess, and Yoshua Bengio.
\newblock Stateful active facilitator: Coordination and environmental
  heterogeneity in cooperative multi-agent reinforcement learning.
\newblock In \emph{The Eleventh International Conference on Learning
  Representations}, 2023.
\newblock URL \url{https://openreview.net/forum?id=B4maZQLLW0_}.

\bibitem[Lu et~al.(2018)Lu, Liu, Dong, Gu, Gama, and Zhang]{Lu_2018}
Jie Lu, Anjin Liu, Fan Dong, Feng Gu, Joao Gama, and Guangquan Zhang.
\newblock Learning under concept drift: A review.
\newblock \emph{IEEE Transactions on Knowledge and Data Engineering}, pp.\
  1–1, 2018.
\newblock ISSN 2326-3865.
\newblock \doi{10.1109/tkde.2018.2876857}.
\newblock URL \url{http://dx.doi.org/10.1109/TKDE.2018.2876857}.

\bibitem[Machado et~al.(2018)Machado, Bellemare, Talvitie, Veness, Hausknecht,
  and Bowling]{machado18arcade}
Marlos~C. Machado, Marc~G. Bellemare, Erik Talvitie, Joel Veness, Matthew~J.
  Hausknecht, and Michael Bowling.
\newblock Revisiting the arcade learning environment: Evaluation protocols and
  open problems for general agents.
\newblock \emph{Journal of Artificial Intelligence Research}, 61:\penalty0
  523--562, 2018.

\bibitem[Micheli et~al.(2023)Micheli, Alonso, and Fleuret]{iris2023}
Vincent Micheli, Eloi Alonso, and Fran{\c{c}}ois Fleuret.
\newblock Transformers are sample-efficient world models.
\newblock In \emph{The Eleventh International Conference on Learning
  Representations}, 2023.
\newblock URL \url{https://openreview.net/forum?id=vhFu1Acb0xb}.

\bibitem[Mnih et~al.(2015)Mnih, Kavukcuoglu, Silver, Rusu, Veness, Bellemare,
  Graves, Riedmiller, Fidjeland, Ostrovski, Petersen, Beattie, Sadik,
  Antonoglou, King, Kumaran, Wierstra, Legg, and Hassabis]{Mnih-V}
Volodymyr Mnih, Koray Kavukcuoglu, David Silver, Andrei~A. Rusu, Joel Veness,
  Marc~G. Bellemare, Alex Graves, Martin Riedmiller, Andreas~K. Fidjeland,
  Georg Ostrovski, Stig Petersen, Charles Beattie, Amir Sadik, Ioannis
  Antonoglou, Helen King, Dharshan Kumaran, Daan Wierstra, Shane Legg, and
  Demis Hassabis.
\newblock Human-level control through deep reinforcement learning.
\newblock \emph{Nature}, 2015.
\newblock \doi{10.1038/nature14236}.
\newblock URL \url{https://doi.org/10.1038/nature14236}.

\bibitem[Nair et~al.(2015)Nair, Srinivasan, Blackwell, Alcicek, Fearon, Maria,
  Panneershelvam, Suleyman, Beattie, Petersen, Legg, Mnih, Kavukcuoglu, and
  Silver]{nair2015massivelyparallelmethodsdeep}
Arun Nair, Praveen Srinivasan, Sam Blackwell, Cagdas Alcicek, Rory Fearon,
  Alessandro~De Maria, Vedavyas Panneershelvam, Mustafa Suleyman, Charles
  Beattie, Stig Petersen, Shane Legg, Volodymyr Mnih, Koray Kavukcuoglu, and
  David Silver.
\newblock Massively parallel methods for deep reinforcement learning, 2015.
\newblock URL \url{https://arxiv.org/abs/1507.04296}.

\bibitem[Osband et~al.(2020)Osband, Doron, Hessel, Aslanides, Sezener, Saraiva,
  McKinney, Lattimore, {Sz}epesv{\'a}ri, Singh, Van~Roy, Sutton, Silver, and
  van Hasselt]{osband2020bsuite}
Ian Osband, Yotam Doron, Matteo Hessel, John Aslanides, Eren Sezener, Andre
  Saraiva, Katrina McKinney, Tor Lattimore, Csaba {Sz}epesv{\'a}ri, Satinder
  Singh, Benjamin Van~Roy, Richard Sutton, David Silver, and Hado van Hasselt.
\newblock Behaviour suite for reinforcement learning.
\newblock In \emph{International Conference on Learning Representations}, 2020.
\newblock URL \url{https://openreview.net/forum?id=rygf-kSYwH}.

\bibitem[Park et~al.(2025)Park, Frans, Eysenbach, and
  Levine]{park2025ogbenchbenchmarkingofflinegoalconditioned}
Seohong Park, Kevin Frans, Benjamin Eysenbach, and Sergey Levine.
\newblock Ogbench: Benchmarking offline goal-conditioned rl, 2025.
\newblock URL \url{https://arxiv.org/abs/2410.20092}.

\bibitem[Paster et~al.(2022)Paster, McIlraith, and Ba]{10.5555/3600270.3603094}
Keiran Paster, Sheila~A. McIlraith, and Jimmy Ba.
\newblock You can't count on luck: why decision transformers and rvs fail in
  stochastic environments.
\newblock In \emph{Proceedings of the 36th International Conference on Neural
  Information Processing Systems}, NIPS '22, Red Hook, NY, USA, 2022. Curran
  Associates Inc.
\newblock ISBN 9781713871088.

\bibitem[Robine et~al.(2023)Robine, H{\"o}ftmann, Uelwer, and
  Harmeling]{robine2023transformerbased}
Jan Robine, Marc H{\"o}ftmann, Tobias Uelwer, and Stefan Harmeling.
\newblock Transformer-based world models are happy with 100k interactions.
\newblock In \emph{The Eleventh International Conference on Learning
  Representations}, 2023.
\newblock URL \url{https://openreview.net/forum?id=TdBaDGCpjly}.

\bibitem[Samvelyan et~al.(2021)Samvelyan, Kirk, Kurin, Parker-Holder, Jiang,
  Hambro, Petroni, Kuttler, Grefenstette, and
  Rockt{\"a}schel]{samvelyan2021minihack}
Mikayel Samvelyan, Robert Kirk, Vitaly Kurin, Jack Parker-Holder, Minqi Jiang,
  Eric Hambro, Fabio Petroni, Heinrich Kuttler, Edward Grefenstette, and Tim
  Rockt{\"a}schel.
\newblock Minihack the planet: A sandbox for open-ended reinforcement learning
  research.
\newblock In \emph{Thirty-fifth Conference on Neural Information Processing
  Systems Datasets and Benchmarks Track (Round 1)}, 2021.
\newblock URL \url{https://openreview.net/forum?id=skFwlyefkWJ}.

\bibitem[Sutton \& Barto(2018)Sutton and Barto]{suttonBarto2018}
Richard~S. Sutton and Andrew~G. Barto.
\newblock \emph{Reinforcement Learning: An Introduction}.
\newblock A Bradford Book, Cambridge, MA, USA, 2018.
\newblock ISBN 0262039249.

\bibitem[Vamplew et~al.(2022)Vamplew, Foale, and Dazeley]{Vamplew2022}
Peter Vamplew, Cameron Foale, and Richard Dazeley.
\newblock The impact of environmental stochasticity on value-based
  multiobjective reinforcement learning.
\newblock \emph{Neural Computing and Applications}, 34\penalty0 (3):\penalty0
  1783--1799, 2022.
\newblock ISSN 1433-3058.
\newblock \doi{10.1007/s00521-021-05859-1}.
\newblock URL \url{https://doi.org/10.1007/s00521-021-05859-1}.

\bibitem[Ye et~al.(2021)Ye, Liu, Kurutach, Abbeel, and
  Gao]{ye2021masteringatarigameslimited}
Weirui Ye, Shaohuai Liu, Thanard Kurutach, Pieter Abbeel, and Yang Gao.
\newblock Mastering atari games with limited data, 2021.
\newblock URL \url{https://arxiv.org/abs/2111.00210}.

\bibitem[Zhang et~al.(2020)Zhang, Chen, Xiao, Li, Liu, Boning, and
  Hsieh]{NEURIPS2020_f0eb6568}
Huan Zhang, Hongge Chen, Chaowei Xiao, Bo~Li, Mingyan Liu, Duane Boning, and
  Cho-Jui Hsieh.
\newblock Robust deep reinforcement learning against adversarial perturbations
  on state observations.
\newblock In H.~Larochelle, M.~Ranzato, R.~Hadsell, M.F. Balcan, and H.~Lin
  (eds.), \emph{Advances in Neural Information Processing Systems}, volume~33,
  pp.\  21024--21037. Curran Associates, Inc., 2020.
\newblock URL
  \url{https://proceedings.neurips.cc/paper_files/paper/2020/file/f0eb6568ea114ba6e293f903c34d7488-Paper.pdf}.

\bibitem[Zhang et~al.(2023)Zhang, Wang, Sun, Yuan, and Huang]{zhang2023storm}
Weipu Zhang, Gang Wang, Jian Sun, Yetian Yuan, and Gao Huang.
\newblock {STORM}: Efficient stochastic transformer based world models for
  reinforcement learning.
\newblock In \emph{Thirty-seventh Conference on Neural Information Processing
  Systems}, 2023.
\newblock URL \url{https://openreview.net/forum?id=WxnrX42rnS}.

\bibitem[Zouitine et~al.(2024)Zouitine, Bertoin, Clavier, Geist, and
  Rachelson]{zouitine2024rrlsrobustreinforcement}
Adil Zouitine, David Bertoin, Pierre Clavier, Matthieu Geist, and Emmanuel
  Rachelson.
\newblock Rrls : Robust reinforcement learning suite, 2024.
\newblock URL \url{https://arxiv.org/abs/2406.08406}.

\end{thebibliography}
\bibliographystyle{iclr2026_conference}

\clearpage
\appendix
\section{Appendix}

\subsection{STORI Implementation}
\label{stori-implementation-appendix}

The STORI framework is built around a sophisticated wrapper-based architecture that introduces various types of uncertainty and partial observability into deterministic Atari environments with a granular control over the modifications.

\subsubsection{Core Architecture and Wrapper System}

\begin{figure}[htbp]
\begin{center}
\includegraphics[width=0.8\textwidth]{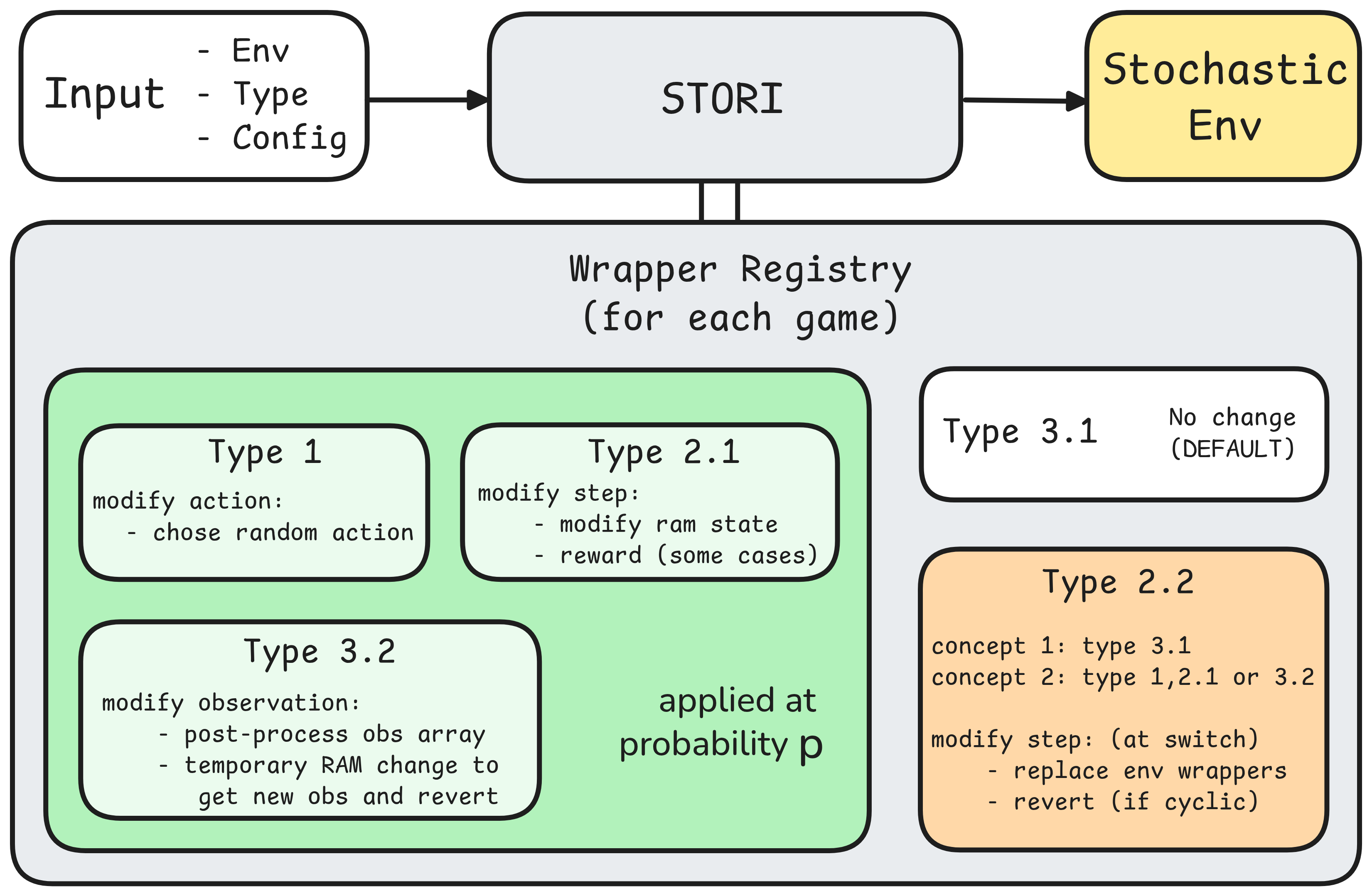}
\end{center}
\caption{STORI Implementation Overview}
\label{fig:STORI-implementation}
\end{figure}

The implementation uses a hierarchical wrapper system built on top of the Atari Learning Environment (ALE). The main `StochasticEnv` class serves as the entry point, which applies different types of wrappers from `wrapper\_registry` of specified environment. The system supports five distinct types, each introducing different forms of stochasticity.
The system is highly configurable through a dictionary-based configuration system. Users can specify probabilities for different stochasticity effects, choose between different modes of operation, and configure temporal parameters for concept drift. The wrapper registry system allows for easy extension and customization of stochasticity types for new games or research requirements.

\subsubsection{Stochasticity Wrappers}
\begin{itemize}
    \item Type 0: This type returns the RAM state of the game (a 1-D numpy array) with state labels as the observation. This implementation is an extension of Atari Annotated RAM Interface \citep{anand2020unsupervisedstaterepresentationlearning}.
    \item Type 1: The `ActionDependentStochasticityWrapper` randomly replaces the agent's intended action with a random action from the action space with a specified probability.
    \item Type 2.1: The `ActionIndependentRandomStochasticityWrapper` implements environment specific random events that occur independently of the agent's actions. These effects are applied probabilistically and create unpredictable environmental changes to which the agent must adapt. Read more about game-specific modifications in section \ref{all_modes}.
    \item Type 2.2: This introduces temporal concept drift where the environment dynamics change over time. The `ActionIndependentConceptDriftWrapper` supports both sudden and cyclic modes between 2 concepts. The concept 1 is the default environment (type 3.1) and concept 2 can be any other environment stochasticity types out of 1, 2.1 and 3.2.  In sudden mode, the environment switches to concept 2 after a fixed number of steps. In cyclic mode, it alternates between the concept 1 and 2 every specified number of steps, creating a challenging environment where the agent must continuously adapt to changing dynamics.
    \item Types 3.1: This stochasticity type returns the default ALE environment without any modifications.
    \item Types 3.2: The `PartialObservationWrapper` introduces partial observability by modifying the agent's observations. The system supports multiple observation modification techniques including cropping (removing portions of the screen), blackout (hiding specific regions), and RAM manipulation (temporarily modifying the game's internal state to get modified observation).

\end{itemize}

In STORI, stochasticity types 1, 2.1, 2.2, and 3.2 are implemented as extensions of Type 3.1 environments. This is because screen-based observations serve as the default, well-studied ALE inputs for various reinforcement learning algorithms, providing a consistent foundation for comparing different types of stochasticity while also allowing for interpretable analysis of agent actions and behaviors.

\subsubsection{Algorithms Additional Details}
\begin{itemize}
    \item DreamerV3: The source implementation and default parameters for Atari100K config used from this code repository (MIT license): \url{https://github.com/NM512/dreamerv3-torch}
    \item STORM: The source implementation and default parameters (except eval num\_episode was set to 100) used from this code repository: \url{https://github.com/weipu-zhang/STORM}
\end{itemize}

\section{Additional Benchmark Details}

\begin{figure}[htbp]
\begin{center}
\includegraphics[width=1\textwidth]{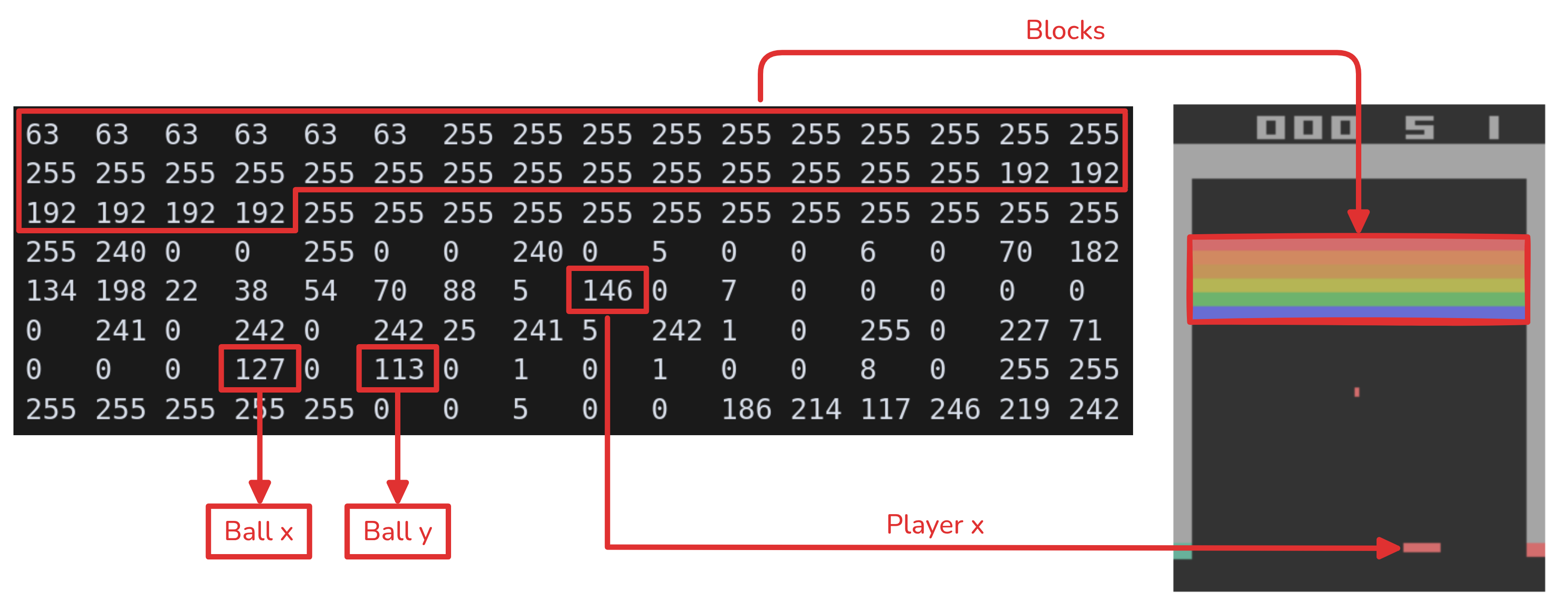}
\end{center}
\caption{The figure shows the RAM state of Atari Breakout on the left and corresponding observation image from the emulator on the right, along with annotations for various state variables like ball position, blocks state etc.}
\label{fig:type0-breakout}
\end{figure}

\clearpage
\subsection{Experiment Stochasticity Modes}
\label{exp-stochastic-modes}

\subsubsection{Stochasticity Modes Used in Breakout Experiments}

\begin{itemize}
    \item Type 1: Random action executed from action space instead of  predicted action with a probability of $0.3$.
    \item Type 2.1: If a block is hit, there is probability of $0.25$ that the hit is not considered and the block is not destroyed thereby returning 0 reward and the ball bounces back.
    \item Type 2.2: Episode starts with default setting (Type 3.1) and after 300 steps into the episode, the dynamics suddenly change to \textit{Type 3.2A}.
    \item Type 3.1: Default Atari Breakout.
    \item Type 3.2A: The ball is only visible is a specific window between the blocks and the paddle and permanently hidden ($p=1.0$) in rest of the space between them.
    \item Type 3.2B: Randomly hide left vertical half of the screen 75\% ($p=0.75$) of the episode.
    \item Type 3.2C: Only a random circular area of the screen is visible every frame ($p=1.0$) similar to what someone will see when walking in a dark room with a torch.
\end{itemize}

\subsubsection{Stochasticity Modes Used in Boxing Experiments}

\begin{itemize}
\item Type 1: Random action executed from action space instead of  predicted action with a probability of $0.3$.
\item Type 2.1: Swaps the color of the enemy and player (character and score) with probability of $0.001$ which results in 6-7 persistent swaps per episode (2 mins boxing round).
\item Type 2.2: Episode starts with default setting (Type 3.1) and after 300 steps into the episode, the dynamics suddenly change to \textit{Type 3.2C}.
\item Type 3.1: Default Atari Boxing.
\item Type 3.2A: Permanently hide ($p=1.0$) scores and game clock.
\item Type 3.2B: Randomly hide right vertical half of the screen 75\% ($p=0.75$) of the episode.
\item Type 3.2C: Randomly hide enemy character 70\% ($p=0.7$) of the episode.
\end{itemize}

\subsubsection{Stochasticity Modes Used in Gopher Experiments}
\begin{itemize}
\item Type 1: Random action executed from action space instead of  predicted action with a probability of $0.3$.
\item Type 2.1: Hole doesn't fill underground below the farmer and the reward is reverted to 0 whenever farmer digs, with probability of $0.3$. 
\item Type 2.2: At the beginning of each episode, the environment is set to the default mode (Type 3.1). Every 600 steps, the dynamics transition \textit{cyclically} between Type 3.2 and the default.
\item Type 3.1: Default Atari Gopher.
\item Type 3.2: Permanently hide ($p=1.0$) underground gopher movement and holes and only hole openings are visible on surface (if any).
\end{itemize}

\subsubsection{Stochasticity Modes Used in BankHeist Experiments}
\begin{itemize}
\item Type 1: With probability $0.3$, a random action is executed from a restricted subset of the action space (0–9) instead of the predicted action. The restriction reduces the frequency of fire-based actions during random sampling, preventing the agent from instantly dying by triggering a bomb it drops on itself.
\item Type 2.1: With probability $0.001$, the robber is unexpectedly teleported to a different city.
\item Type 2.2: At the beginning of each episode, the environment is set to the default mode (Type 3.1). Every 600 steps, the dynamics transition \textit{cyclically} between Type 3.2 and the default.
\item Type 3.1: Default Atari BankHeist.
\item Type 3.2: Randomly hide city blocks 75\% ($p=0.75$) of the frames.
\end{itemize}

\clearpage
\subsection{Learning Curves For Different Stochasticity Types}

\begin{figure}[htbp]
\begin{center}
\includegraphics[width=1\textwidth]{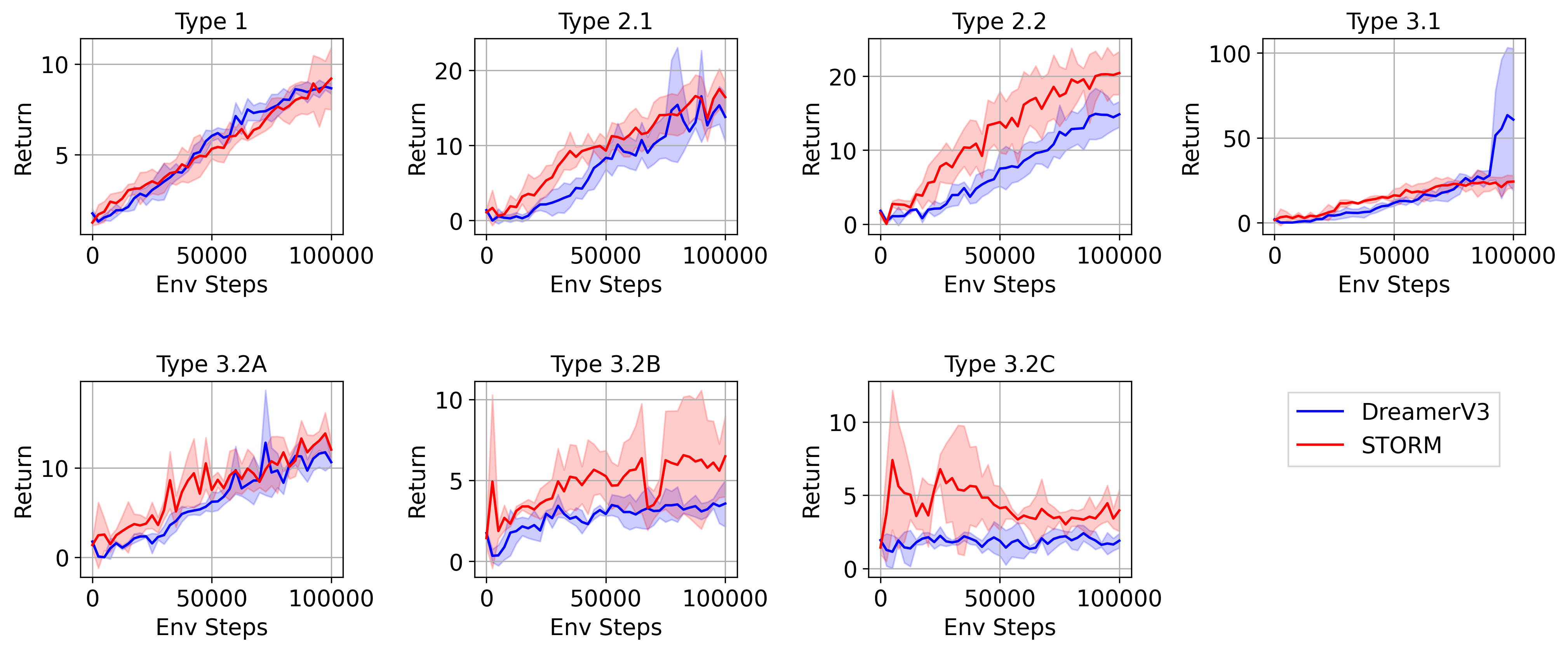}
\end{center}
\caption{Breakout - learning curves}
\label{fig:breakout_training_curves}
\end{figure}

\begin{figure}[htbp]
\begin{center}
\includegraphics[width=1\textwidth]{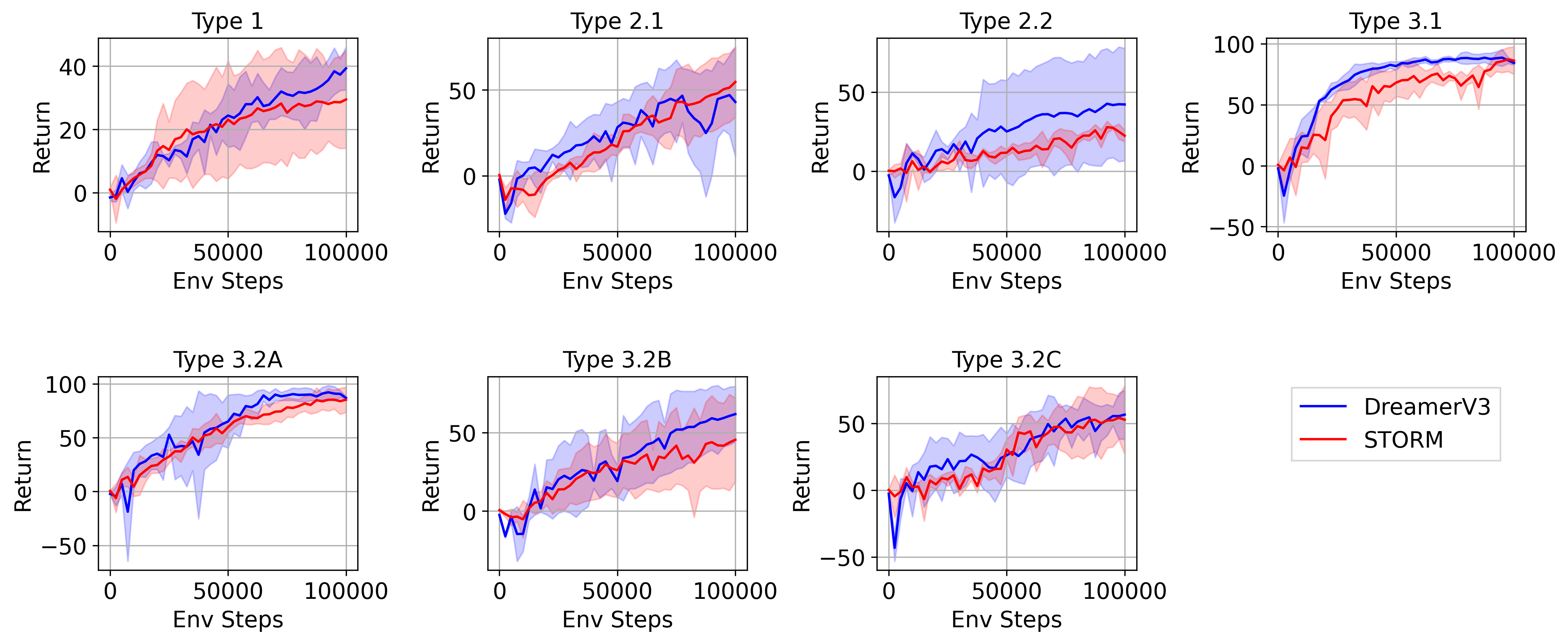}
\end{center}
\caption{Boxing - learning curves}
\label{fig:boxing_training_curves}
\end{figure}

\begin{figure}[htbp]
\begin{center}
\includegraphics[width=1\textwidth]{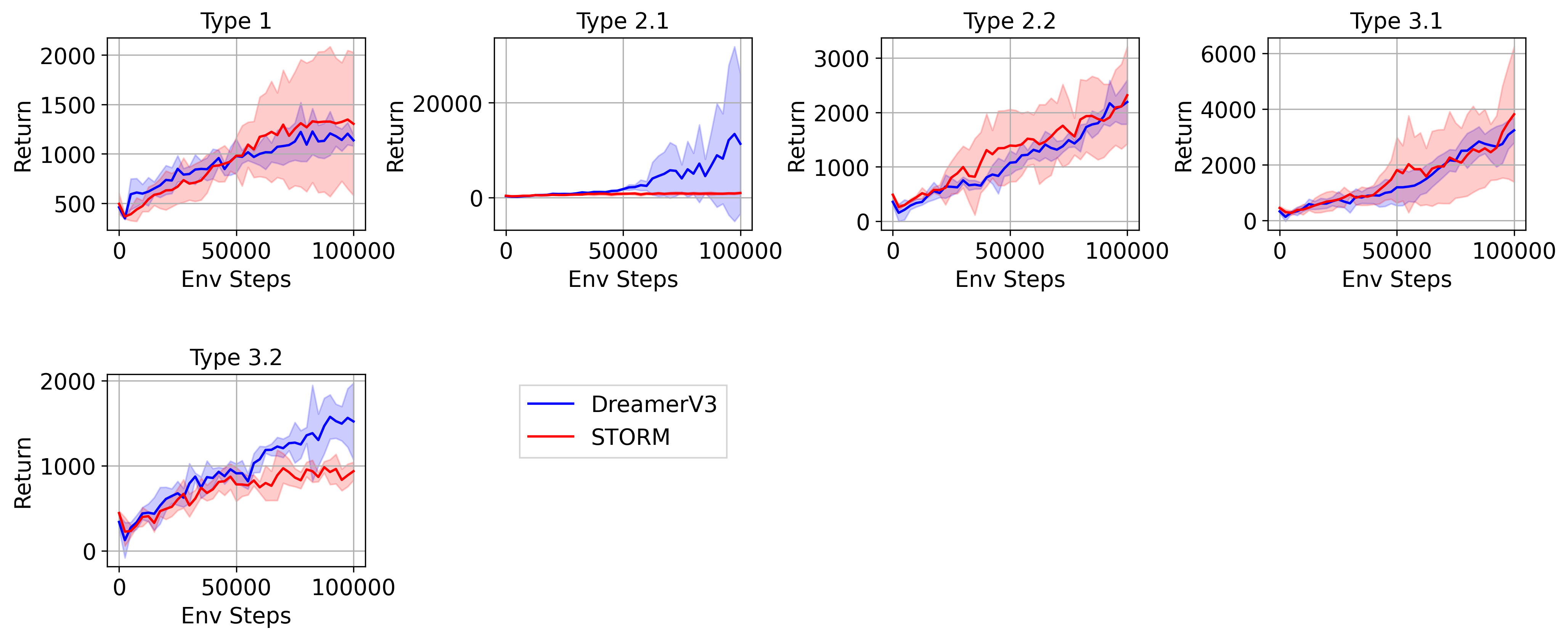}
\end{center}
\caption{Gopher - learning curves}
\label{fig:gopher_training_curves}
\end{figure}

\begin{figure}[htbp]
\begin{center}
\includegraphics[width=1\textwidth]{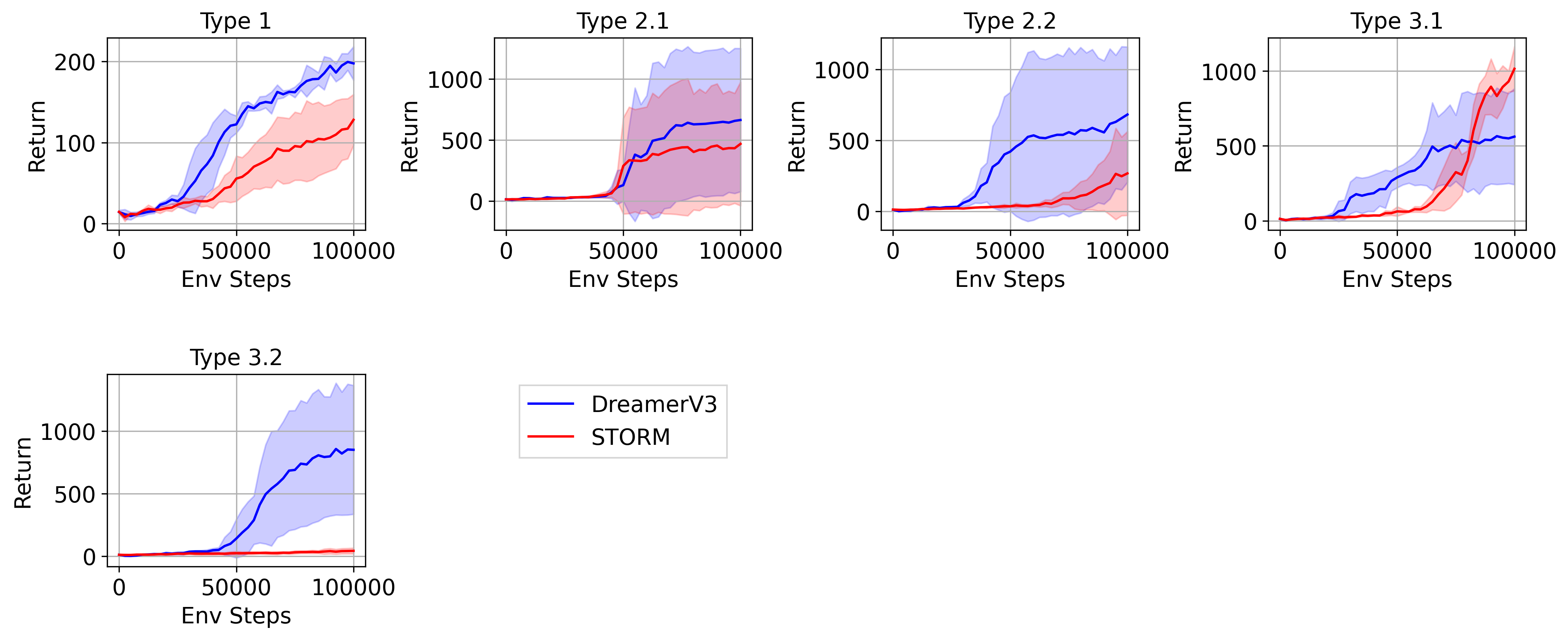}
\end{center}
\caption{BankHeist - learning curves}
\label{fig:bankheist_training_curves}
\end{figure}

Figures \ref{fig:breakout_training_curves}, \ref{fig:boxing_training_curves}, \ref{fig:gopher_training_curves}, and \ref{fig:bankheist_training_curves} illustrate the learning curves on Breakout, Boxing Gopher and BankHeist respectively, depicting the average evaluation return as a function of training steps up to 100K, for DreamerV3 and STORM.

\clearpage
\subsection{Overall Results for Evaluation Return}

\begin{table}[htbp]
\caption{Overall Results for Evaluation Return}
\label{all_results}
\begin{center}
\begin{tabular}{clrr}
\multicolumn{1}{c}{\bf GAME NAME}  &
\multicolumn{1}{p{3cm}}{\bf STOCHASTICITY TYPE} &
\multicolumn{1}{r}{\bf DREAMERV3} &
\multicolumn{1}{r}{\bf STORM} 
\\ \hline
\noalign{\vskip 2pt} % adds vertical padding after the cline
\multirow{7}{*}{Breakout} & 1    & $8.67\pm0.30$ & $9.20\pm1.72$ \\
\noalign{\vskip 2pt} % adds vertical padding before the cline
\cline{2-4}
\noalign{\vskip 2pt} % adds vertical padding after the cline
                          & 2.1  & $13.80\pm3.26$ & $16.44\pm2.03$ \\
                          & 2.2  & $14.86\pm1.74$ & $20.45\pm2.92$ \\
\noalign{\vskip 2pt} % adds vertical padding before the cline
\cline{2-4}
\noalign{\vskip 2pt} % adds vertical padding after the cline
                          & 3.1 (Default Baseline)  & $60.71\pm41.89$ & $24.17\pm3.55$ \\
                          & 3.2A & $10.65\pm0.41$ & $12.05\pm0.89$ \\
                          & 3.2B & $3.57\pm1.43$ & $6.48\pm2.50$ \\
                          & 3.2C & $1.89\pm0.49$ & $3.96\pm1.42$ \\
\hline
\noalign{\vskip 2pt} % adds vertical padding after the cline
\multirow{7}{*}{Boxing} & 1    & $39.32\pm6.79$ & $29.48\pm15.41$ \\
\noalign{\vskip 2pt} % adds vertical padding before the cline
\cline{2-4}
\noalign{\vskip 2pt} % adds vertical padding after the cline
                          & 2.1  & $43.00\pm31.98$ & $54.69\pm20.44$ \\
                          & 2.2  & $42.21\pm35.54$ & $22.44\pm3.54$ \\
\noalign{\vskip 2pt} % adds vertical padding before the cline
\cline{2-4}
\noalign{\vskip 2pt} % adds vertical padding after the cline
                          & 3.1 (Default Baseline)  & $84.22\pm1.68$ & $86.18\pm11.29$ \\
                          & 3.2A & $86.90\pm1.33$ & $85.22\pm11.77$ \\
                          & 3.2B & $61.74\pm17.71$ & $45.41\pm26.56$ \\
                          & 3.2C & $56.52\pm18.45$ & $52.66\pm25.68$ \\
\hline
\noalign{\vskip 2pt} % adds vertical padding after the cline
\multirow{5}{*}{Gopher} & 1    & $1137.00\pm56.98$ & $1303.53\pm724.76$ \\
\noalign{\vskip 2pt} % adds vertical padding before the cline
\cline{2-4}
\noalign{\vskip 2pt} % adds vertical padding after the cline
                          & 2.1  & $11333.53\pm14761.12$ & $950.67\pm188.37$ \\
                          & 2.2  & $2190.87\pm407.24$ & $2315.40\pm893.32$ \\
\noalign{\vskip 2pt} % adds vertical padding before the cline
\cline{2-4}
\noalign{\vskip 2pt} % adds vertical padding after the cline
                          & 3.1 (Default Baseline)  & $3235.27\pm443.51$ & $3811.67\pm2431.85$ \\
                          & 3.2 & $1521.13\pm451.45$ & $936.40\pm106.83$ \\
\hline
\noalign{\vskip 2pt} % adds vertical padding after the cline
\multirow{5}{*}{BankHeist} & 1    & $197.63\pm20.76$ & $128.03\pm31.41$ \\
\noalign{\vskip 2pt} % adds vertical padding before the cline
\cline{2-4}
\noalign{\vskip 2pt} % adds vertical padding after the cline
                          & 2.1  & $663.60\pm587.85$ & $467.80\pm507.74$ \\
                          & 2.2  & $682.67\pm476.90$ & $267.70\pm295.86$ \\
\noalign{\vskip 2pt} % adds vertical padding before the cline
\cline{2-4}
\noalign{\vskip 2pt} % adds vertical padding after the cline
                          & 3.1 (Default Baseline)  & $562.30\pm320.74$ & $1015.73\pm148.43$ \\
                          & 3.2 & $849.80\pm514.15$ & $43.10\pm22.85$ \\
\hline
\end{tabular}
\end{center}
\end{table}

\clearpage
\subsection{Additional Details: Type 3.2 Error Analysis}
\label{sec:appendix_type_3_2}

\begin{figure}[htbp]
\begin{center}
\includegraphics[width=0.8\textwidth]{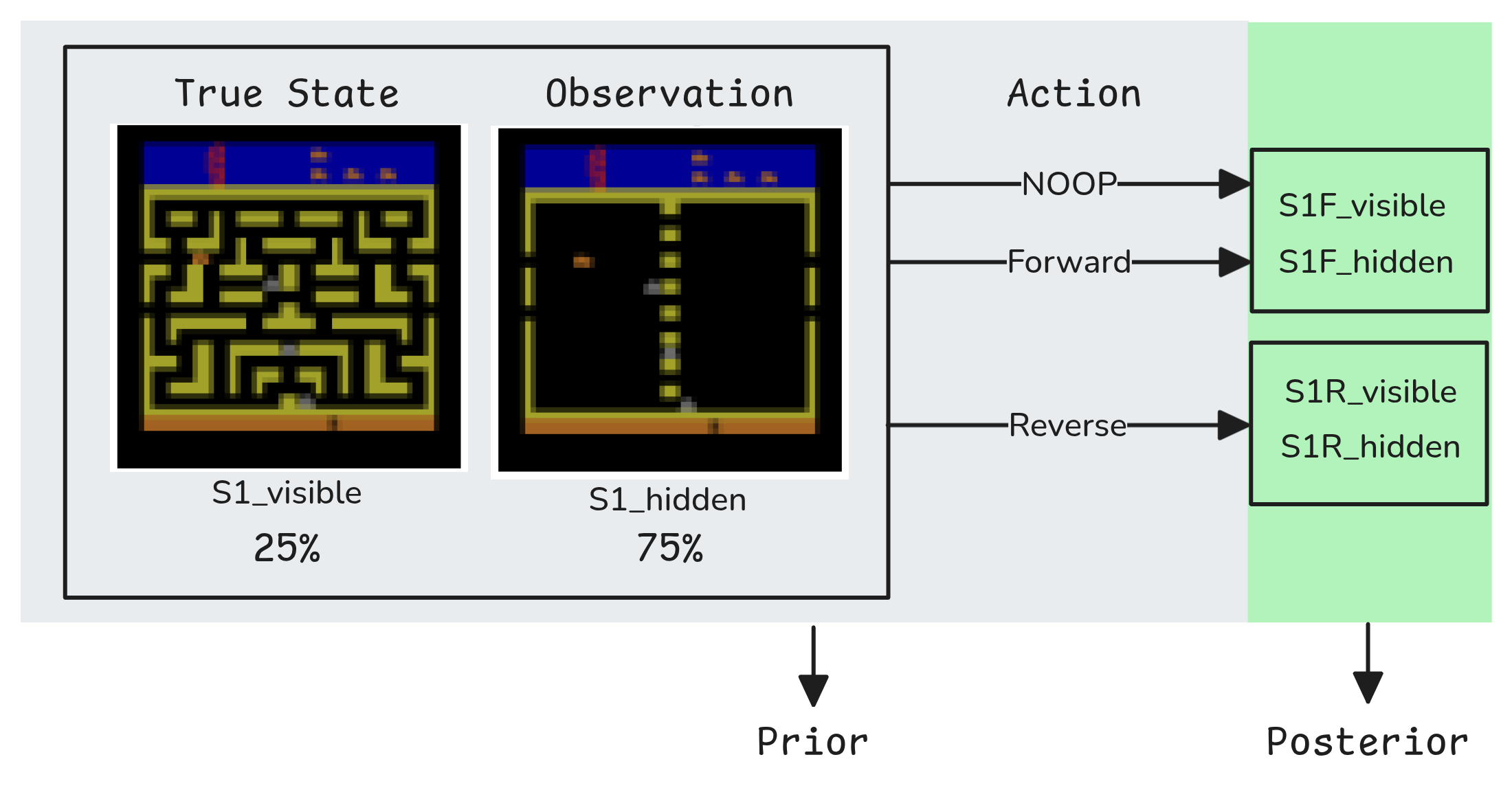}
\end{center}
\caption{Detailed results from the BankHeist Type 3.2 partial observability probe. Each row corresponds to one of six carefully selected cases, showing the start state visibility, next state visibility, negative log-likelihood (NLL), and KL divergence for both DreamerV3 and STORM.}
\label{fig:type_3_error analysis}
\end{figure}

\begin{table}[htbp]
\centering
\caption{Analysis on BankHeist (Type 3.2) - DreamerV3}
\label{tab:DreamerV3_bankheist_type3}
\begin{tabular}{cccccccc}
\hline
\noalign{\vskip 2pt}
\multirow{3}{*}{\textbf{Prior Obs}}& \multirow{3}{*}{\textbf{Metrics}}& \multicolumn{6}{c}{\textbf{Action | Posterior Observations (Visible/Hidden)}}\\
 & & NOOP (0)& Forward (3)& Reverse (4)& NOOP (0)& Forward (3)&Reverse (4)\\
& & S1F\_visible& S1F\_visible & S1R\_visible &S1F\_hidden& S1F\_hidden & S1R\_hidden \\ \hline
\noalign{\vskip 2pt}
\multirow{2}{*}{S1\_visible} & - log d& 41.72& 38.43& 44.73& 19.13& 21.56& 29.81\\ \cline{2-8}
\noalign{\vskip 2pt}
& KL div & 25.52& 20.13& 24.71& 6.40& 7.17& 12.78\\ \hline
\noalign{\vskip 2pt}
\multirow{2}{*}{S1\_hidden} & - log d& 41.36& 38.74& 45.14& 25.15& 22.63& 29.25\\ \cline{2-8}
\noalign{\vskip 2pt}
& KL div & 20.99& 20.47& 24.57& 10.72& 7.05& 11.81\\ \hline
\end{tabular}
\end{table}

\begin{table}[htbp]
\centering
\caption{Analysis on BankHeist (Type 3.2) - STORM}
\label{tab:STORM_bankheist_type3}
\begin{tabular}{cccccccc}
\hline
\noalign{\vskip 2pt}
\multirow{3}{*}{\textbf{Prior Obs}}& \multirow{3}{*}{\textbf{Metrics}}& \multicolumn{6}{c}{\textbf{Action | Posterior Observations (Visible/Hidden)}}\\
 & & NOOP (0)& Forward (3)& Reverse (4)& NOOP (0)& Forward (3)&Reverse (4)\\
& & S1\_visible & S1F\_visible & S1R\_visible &S1\_hidden & S1F\_hidden & S1R\_hidden \\ \hline
\noalign{\vskip 2pt}
\multirow{2}{*}{S1\_visible} & - log p& 33.53& 21.72& 28.65& 19.81& 11.29& 21.86\\ \cline{2-8}
\noalign{\vskip 2pt}
& KL div & 115.63& 116.00& 114.13& 114.94& 115.47& 113.79\\ \hline
\noalign{\vskip 2pt}
\multirow{2}{*}{S1\_hidden} & - log p& 28.60& 18.50& 22.58& 18.43& 12.61& 16.21\\ \cline{2-8}
\noalign{\vskip 2pt}
& KL div & 115.55& 115.49& 115.05& 115.03& 115.19& 114.73\\ \hline
\end{tabular}
\end{table}

\clearpage
\subsection{Compute Resources Used}
\label{compute_resource}
The experiment runs were executed in several types of GPUs like A40, A100 and H100 depending on availability. Each node atleast had 32 vCPU and 50GB RAM. On GPUs with large memory, mulitple runs were executed.

DreamerV3 and STORM took around 24 hours and 12 hours respectively per run (training \& evaluation) per seed when running on single GPU.

\section{Information-Theoretic Levers}
Information-theoretic measures provide quantitative levers to diagnose how stochasticity affects learning and planning:

\paragraph{Action channel capacity.} For action-dependent noise, controllability is reduced. The effective capacity is measured by $I(A;\tilde{A} \mid S)$, which quantifies how much of the intended action $A$ survives corruption into the executed action $\tilde{A}$.

\paragraph{Predictive information of dynamics.} For action-independent randomness, the predictive structure is measured by $I((S_t,A_t); S_{t+1})$, reflecting how much the next state depends on the current state-action pair. Under drift, temporal changes in this quantity indicate shifts in environment regularity.

\paragraph{Representation sufficiency.} A latent $Z_t$ should act as a sufficient statistic for planning. Ideally, $I(Z_t; S_t)$ is maximized, while $I(Z_t; O_t)$ remains bounded, ensuring that $Z_t$ captures hidden states rather than surface-level noise, consistent with bisimulation invariance.

\paragraph{Aliasing quantification.} In partially observable settings, the observation-state information gap can be written as $I(S_t;O_t) - I(S_t;O_t \mid A_t)$, capturing residual uncertainty after conditioning on actions. This disentangles sensor noise from genuine state ambiguity.

\paragraph{Risk-sensitive planning.} Robust planning can be viewed through an information lens: risk-sensitive objectives such as Conditional Value at Risk (CVaR) optimize not the mean return but lower quantiles, effectively re-weighting information from rare but catastrophic outcomes.

\clearpage
\section{All Implemented Stochasticity Modes}
\label{all_modes}

We define stochasticity modes along four Atari environments (Breakout, Boxing, Gopher, BankHeist), and the set of cropping modes are common to all games.

\subsection*{Common Cropping Modes (All Games)}
\begin{itemize}
  \item Mode 0: No crop
  \item Mode 1: Left --- Crop the left half of the observation
  \item Mode 2: Right --- Crop the right half of the observation
  \item Mode 3: Top --- Crop the top half of the observation
  \item Mode 4: Bottom --- Crop the bottom half of the observation
  \item Mode 5: Random circular mask --- Randomly mask a circular region of the observation
\end{itemize}

\subsection*{Breakout}
\textbf{Action-independent random}
\begin{itemize}
  \item 0: none
  \item 1: block hit cancel (reward unchanged)
  \item 2: block hit cancel (reward set to 0)
  \item 3: regenerate a randomly chosen hit block
\end{itemize}

\textbf{Partial observation (blackout)}
\begin{itemize}
  \item 0: none
  \item 1: all
  \item 2: blocks
  \item 3: paddle
  \item 4: score
  \item 5: ball\_missing\_top
  \item 6: ball\_missing\_middle
  \item 7: ball\_missing\_bottom
  \item 8: blocks\_and\_paddle
  \item 9: blocks\_and\_score
  \item 10: ball\_missing\_top\_and\_bottom
  \item 11: ball\_missing\_all
\end{itemize}

\textbf{Partial observation - RAM modification}
\begin{itemize}
  \item 0: none
  \item 1: nus\_pattern (blocks RAM)
  \item 2: ball\_hidden
\end{itemize}

\subsection*{Boxing}
\textbf{Action-independent random}
\begin{itemize}
  \item 0: none
  \item 1: colorflip (swap player/enemy colors)
  \item 2: hit cancel (revert score; reward set to 0)
  \item 3: displace to corners (swap player/enemy positions)
\end{itemize}

\textbf{Partial observation (blackout)}
\begin{itemize}
  \item 0: none
  \item 1: all
  \item 2: left boxing ring
  \item 3: right boxing ring
  \item 4: full boxing ring
  \item 5: enemy score
  \item 6: player score
  \item 7: enemy+player score
  \item 8: clock
  \item 9: enemy+player score+clock
\end{itemize}

\textbf{Partial observation - RAM modification}
\begin{itemize}
  \item 0: none
  \item 1: hide boxing ring
  \item 2: hide enemy
  \item 3: hide player
\end{itemize}

\subsection*{Gopher}
\textbf{Action-independent random}
\begin{itemize}
  \item 0: none
  \item 1: hole doesn’t close (fill cancel; reward unchanged)
  \item 2: hole doesn’t close (fill cancel; reward set to 0)
  \item 3: randomly remove one visible carrot (once per reset)
\end{itemize}

\textbf{Partial observation (blackout)}
\begin{itemize}
  \item 0: none
  \item 1: all
  \item 2: gopher attack (both sides)
  \item 3: left gopher attack
  \item 4: right gopher attack
  \item 5: underground full (before-dug color)
  \item 6: underground full offset (before-dug color)
  \item 7: underground row 0 (before-dug)
  \item 8: underground row 0 (dug color)
  \item 9: underground row 1 (before-dug)
  \item 10: underground row 1 (dug color)
  \item 11: underground row 2 (before-dug)
  \item 12: underground row 2 (dug color)
  \item 13: underground row 3 (before-dug)
  \item 14: underground row 3 (dug color)
  \item 15: farmer (full)
  \item 16: farmer below nose
  \item 17: duck fly
  \item 18: score
\end{itemize}

\textbf{Partial observation - RAM modification}
\begin{itemize}
  \item 0: none
  \item 1: hide left carrot
  \item 2: hide middle carrot
  \item 3: hide right carrot
  \item 4: hide all carrots
  \item 5: hide seed
\end{itemize}

\subsection*{BankHeist}
\textbf{Action-independent random}
\begin{itemize}
  \item 0: none
  \item 1: dropped bomb is a dud
  \item 2: fuel leaks (per city, once per episode)
  \item 3: switch city mid-way (teleport)
  \item 4: bank empty (reward suppressed when bank$\rightarrow$police transition detected)
\end{itemize}

\textbf{Partial observation (blackout)}
\begin{itemize}
  \item 0: none
  \item 1: all
  \item 2: city walls (all)
  \item 3: top city wall
  \item 4: left city wall
  \item 5: bottom city wall
  \item 6: right city wall
  \item 7: left and right city walls together
  \item 8: fuel region
  \item 9: lives region
  \item 10: score region
\end{itemize}

\textbf{Partial observation - RAM modification}
\begin{itemize}
  \item 0: none
  \item 1: hide robber’s car
  \item 2: hide change in fuel (always full)
  \item 3: hide city blocks
  \item 4: blend city blocks and wall (background color)
  \item 5: hide banks (when currently a bank)
  \item 6: hide police (when currently police)
\end{itemize}

\subsection*{Concept Drift Usage}
All \textit{partial observation}, \textit{action-independent}, and \textit{action-dependent} modes can also be used as a \textbf{second concept} in a concept drift setting, enabling controlled evaluation of robustness to non-stationary environments.

\clearpage
\section{The Use of Large Language Models (LLMs)}

We made use of large language models (LLMs) to assist with selected aspects of this work. Specifically, LLMs were employed to improve the clarity and flow of writing, to summarize and condense long paragraphs during manuscript preparation, and to generate code snippets for repetitive components of the implementation. All outputs from the LLMs were carefully reviewed, validated, and edited by the authors to ensure accuracy and correctness.

\end{document}